\documentclass{article}
\PassOptionsToPackage{numbers, compress}{natbib}

\usepackage[main, final]{neurips_2026}
\usepackage{amsthm}
\usepackage{adjustbox}
\usepackage{amssymb}
\usepackage{placeins}
\newtheorem{proposition}{Proposition}
\newtheorem*{remark}{Remark}

\usepackage[utf8]{inputenc} 
\usepackage[T1]{fontenc}    
\usepackage{hyperref}       
\usepackage{url}            
\usepackage{booktabs}       
\usepackage{amsfonts}       
\usepackage{nicefrac}       
\usepackage{microtype}      
\usepackage{xcolor}         
\usepackage{microtype}
\usepackage{graphicx}
\usepackage{subcaption}
\usepackage{booktabs} 
\usepackage{multirow}
\usepackage{booktabs}
\usepackage{multirow}
\usepackage{xcolor}
\usepackage{graphicx}          
\usepackage[table]{xcolor}
\usepackage{booktabs,multirow,graphicx,rotating}

\newcommand{\best}[1]{\cellcolor{red!18}\textbf{#1}}
\newcommand{\second}[1]{\cellcolor{blue!15}\textbf{#1}}
\newcommand{\capbox}[2]{{\setlength{\fboxsep}{1pt}\colorbox{#1}{\textbf{#2}}}}
\newcommand{\capbest}[1]{\capbox{red!18}{#1}}
\newcommand{\capsecond}[1]{\capbox{blue!15}{#1}}

\newcommand{\ours}{\texttt{EntroPE}}
\newcommand{\oursbold}{\textbf{\texttt{EntroPE}}}
\newcommand{\tabstyle}{\scriptsize
  \setlength{\tabcolsep}{4pt}\renewcommand{\arraystretch}{1.05}}
\usepackage{wrapfig}
\usepackage{hyperref}

\usepackage{amsmath}
\usepackage{amssymb}
\usepackage{mathtools}
\usepackage{amsthm}

\usepackage[textsize=tiny]{todonotes}
\usepackage[capitalize,noabbrev]{cleveref}

\title{Entropy Guided Dynamic Patch Segmentation for Time Series Transformers}

\author{%
Sachith Abeywickrama\textsuperscript{1,2}
\quad
Emadeldeen Eldele\textsuperscript{3}\thanks{Corresponding author. \texttt{emadeldeenahmed.eldele@ku.ac.ae}}
\quad
Min Wu\textsuperscript{2}
\quad
Xiaoli Li\textsuperscript{4}
\quad
Chau Yuen\textsuperscript{1}
\\[1.5mm]
\textsuperscript{1}Nanyang Technological University, Singapore
\\
\textsuperscript{2}Institute of Advanced Intelligence and Computing (IAIC), A*STAR, Singapore
\\
\textsuperscript{3}Department of Computer Science, Khalifa University, UAE
\\
\textsuperscript{4}Singapore University of Technology and Design, Singapore
}

\begin{document}

\maketitle

\begin{abstract}
Patch-based transformers have emerged as efficient, improved architectures for long-horizon time series modeling. Yet existing approaches rely on \textit{temporally-agnostic} patch construction, in which arbitrary starting positions and fixed lengths fracture temporal coherence by splitting natural transitions across patch boundaries. This naive segmentation often disrupts short-term dependencies and weakens representation learning. We propose a novel \textbf{Entro}py-Guided Dynamic \textbf{P}atch \textbf{E}ncoder (\texttt{EntroPE}), as a temporally informed framework that dynamically detects transition points via conditional entropy and dynamically places patch boundaries. This preserves temporal structure while retaining the computational benefits of patching. \texttt{EntroPE} consists of 2 key modules, namely an \textbf{Entropy-Guided Dynamic Patcher (EDP)} that applies information-theoretic criteria to locate natural temporal shifts and determine patch boundaries, and an \textbf{Adaptive Patch Encoder (APE)} that employs pooling and cross-attention to capture intra-patch dependencies and produce fixed-size latent representations. Extensive experiments on long-term forecasting, classification, and anomaly detection demonstrate that the proposed method improves both accuracy and efficiency, establishing entropy-guided dynamic patching as a promising new paradigm for time series modeling. 
Code is available at: \url{https://github.com/Sachithx/EntroPE}.
\end{abstract}

\section{Introduction}

Time series analysis is fundamental to numerous critical applications, including electricity load forecasting \citep{gasparin2022deep}, financial market prediction \citep{fischer2018deep,wickramasinghe2025smart}, healthcare monitoring \citep{reis2003time}, and environmental surveillance \citep{smith1989extreme}. However, accurately modeling time-series data remains challenging due to inherent noise, complex temporal dependencies, and irregular patterns across multiple time horizons \citep{lim2021time, torres2021deep}. The transformer architecture \citep{vaswani2017attention} has revolutionized language and sequence modeling across domains through self-attention mechanisms, leading to significant advances in time series analysis. Despite their promise in learning long-term temporal relationships, time series transformers remain suboptimal due to unique complexities such as non-stationarity, seasonality, and varying temporal granularities \citep{wen2023transformers, zeng2023transformers}.

Recent advances in time series transformers include Informer \citep{zhou2021informer}, Autoformer \citep{wu2021autoformer}, and FEDformer \citep{zhou2022fedformer}, which address long-sequence modeling challenges through sparse attention mechanisms, decomposition techniques, and improved positional encodings while operating on point-wise inputs. Triformer \citep{Triformer} and PatchTST \citep{patchtst2023} introduced a patch-based input representation by dividing raw sequences into fixed-length patches, achieving substantial improvements in computational efficiency and performance. This demonstrated the critical impact of input tokenization strategies on transformer performance in time series. However, subsequent approaches have predominantly adopted similar \textit{temporally agnostic} patching schemes, which may not fully capture the temporal coherence and statistical properties inherent in time-series data. 

\begin{wrapfigure}{r}{0.5\textwidth}
  \centering
  \vspace{-12pt}
  \captionsetup{width=0.9\linewidth}
  \includegraphics[width=0.45\textwidth]{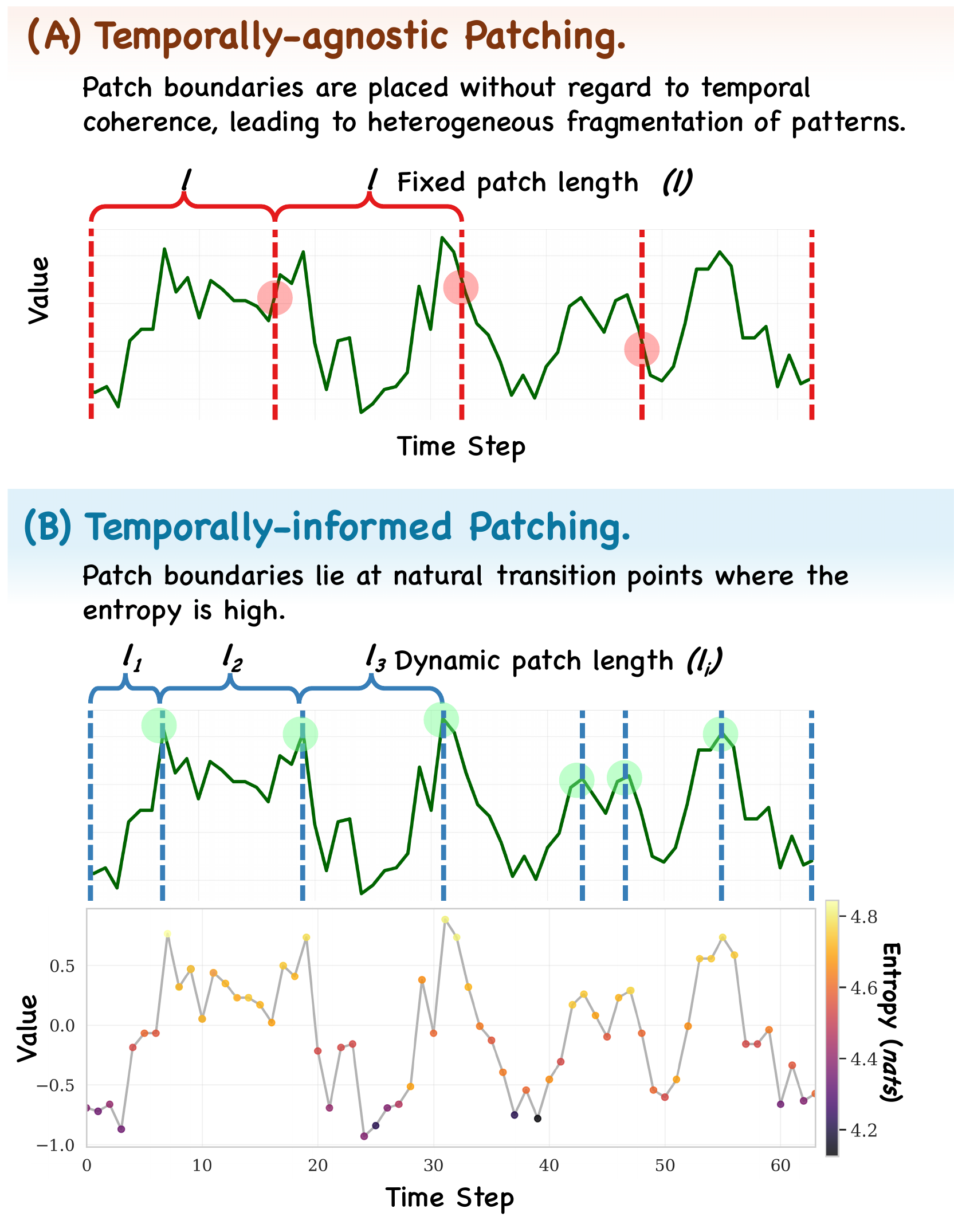}
  \caption{\textbf{Comparison of patching strategies.} (a) Temporally-agnostic fixed-length patching places boundaries at arbitrary positions, fragmenting coherent temporal patterns. (b) Our temporally informed approach places boundaries at entropy peaks, where the conditional entropy scale indicates natural transition points, thereby preserving temporal coherence within patches.}
    \label{fig:intro}
\end{wrapfigure}
These patching strategies treat time as temporally homogeneous, partitioning sequences at arbitrary points without regard to the underlying temporal structure. \textit{Temporally-agnostic} patching introduces 2 fundamental challenges. \textbf{(1)} They create inconsistent input representations between training and inference phases. Training observes patches starting at diverse temporal offsets, which result in patches that begin at various temporal positions relative to underlying patterns (e.g., a patch might start mid-trend in one sample but at a trend beginning in another). However, during inference, patches are extracted from predetermined positions in the input sequence, creating a systematic distribution shift in patch configurations. \textbf{(2)} Further, predetermined patch boundaries arbitrarily segment coherent temporal structures, ignoring frequent underlying dynamics (see Figure \ref{fig:intro}(\textbf{A})). For example, a frequently occurring gradual trend change or seasonal transition may be split across multiple patches, breaking the natural temporal dependencies that are crucial for accurate pattern recognition.

These challenges directly impact model performance. Fragmented temporal patterns create incomplete intra-patch representations, separating semantically related time points across patches and processing them independently. Additionally, the train-inference mismatch reduces the model's ability to generalize, as it must extrapolate to patch configurations with different statistical properties than those encountered during training. This is particularly detrimental for capturing rapid transitions and fine-grained temporal dynamics that require consistent temporal context.

We argue that patch boundaries should align with points of elevated \textbf{predictive uncertainty}, where the past provides insufficient information to determine the future. Formally, such moments are characterized by high conditional entropy of the next observation given historical context. Segmenting time at these entropy peaks localizes regime changes at patch boundaries while yielding patches whose interiors remain predictively coherent. Motivated by this principle, we introduce \texttt{EntroPE}, an \textbf{Entro}py-Guided Dynamic \textbf{P}atch \textbf{E}ncoder, which integrates \textit{uncertainty-aware} boundary detection as a core architectural component. Unlike prior approaches that fix patch length or rely on heuristic change-point detectors, \texttt{EntroPE} derives segmentation from information-theoretic predictive structure, ensuring patch boundaries align with natural temporal transitions (Figure~\ref{fig:intro}(\textbf{B})). We then employ adaptive encoding mechanisms to process these variable-length patches while preserving intra-patch dependencies. This temporally-informed approach addresses both the train-inference mismatch and boundary fragmentation issues inherent in \textit{temporally-agnostic} patching methods. The main contributions of this paper are:
\begin{itemize} 
    \item We propose \texttt{EntroPE}, an information-theoretic framework for dynamic patching in time series transformers. It isolates conditional predictive entropy as a causal boundary signal for time-series patching and shows it outperforms existing alternative segmentation criteria. 
    \item We quantitatively show that conditional-entropy-induced boundaries produce 2.8$\times$ more internally homogeneous patches than fixed patching across 6 datasets, with 97.2\% of detected boundaries coinciding with high-uncertainty positions.
    \item We show that this segmentation principle generalizes across long-term forecasting, classification, and anomaly detection, with efficient computational overhead from the lightweight entropy model.
\end{itemize}

\section{Related Work}

\paragraph{Time-Series Transformers and Patch Representations.} Informer \citep{zhou2021informer}, Autoformer \citep{wu2021autoformer}, and FEDformer \citep{zhou2022fedformer} improve long-sequence forecasting through sparse attention, decomposition, and frequency-domain modeling, respectively. Patch representations offer a complementary approach by grouping observations into local segments. Following patch-based modeling in vision \citep{dosovitskiy2021image}, Triformer \citep{Triformer} and PatchTST \citep{patchtst2023} demonstrate the benefits of temporal patching for efficient forecasting. Subsequent work explores cross-variable dependencies \citep{zhang2023crossformer,xue2024card}, patch-independent self-supervised learning \citep{lee2024learning}, and convolutional or MLP-based patch processing \citep{stitsyuk2025xpatch,tang2025unlocking}.
 
\paragraph{Multi-Scale and Structured Representations.} Several methods enrich patch representations across temporal scales. MSPatch \citep{MSPatch} combines multiple patch resolutions, while Pathformer \citep{chen2024pathformer} adaptively routes computation across multi-scale Transformer branches. MOIRAI \citep{woo2024unified} uses multiple patch-size projections to accommodate different sampling frequencies. Rough Transformers \citep{rough_trans} instead use multi-view path signatures to encode local and global temporal structure. These approaches illustrate how the scale and representation of temporal segments influence modeling efficiency and accuracy.
 
\paragraph{Adaptive Patching and Segmentation.} Adaptation can operate on patch representations, selection, or boundaries. AdaPatch \citep{liu2025adapatch} addresses non-stationarity through patch-level normalization and reconstruction, while HDMixer \citep{huang2024hdmixer} learns extendable patches. APN \citep{liu2025rethinking} learns time-aware patch boundaries and aggregation weights for irregular observations. SRSNet \citep{wu2025enhancing} selects and reassembles informative candidate patches, and TimeMosaic \citep{ding2025timemosaic} adapts patch granularity to temporal heterogeneity. TimeSqueeze \citep{ankireddy2026timesqueeze} combines a state-space encoder with dynamic segmentation, placing boundaries where consecutive-value changes exceed a threshold scaled by local signal power. ReinPatch \citep{wu2026dynamic} learns discrete patch-boundary decisions through reinforcement learning, jointly optimizing a patching policy and the downstream sequence model using Group Relative Policy Gradient under a minimum compression constraint. These methods provide mechanisms for adapting temporal resolution; \texttt{EntroPE} investigates learned predictive uncertainty as the boundary-selection signal.

\paragraph{Entropy-Guided Boundary Selection.} The Byte Latent Transformer (BLT) \citep{pagnoni2025byte} uses next-byte predictive entropy from a small autoregressive model to form variable-length patches for language modeling. In time-series forecasting, EAPformer \citep{ling2025eapformer} uses temporal entropy to adjust patch boundaries and combines this segmentation with multivariate dependency modeling. \texttt{EntroPE} adapts conditional predictive entropy to continuous time series using a separately trained, frozen autoregressive estimator. The entropy estimator and patch encoder receive quantized input tokens, and the forecasting head produces continuous predictions. To examine the contribution of the boundary criterion, we compare learned predictive entropy with empirical window entropy and alternative boundary signals under a matched downstream architecture.

\section{Preliminaries}
\subsection{Problem Formulation}

\begin{figure*}[t]
    \centering
    \includegraphics[width=0.85\linewidth]{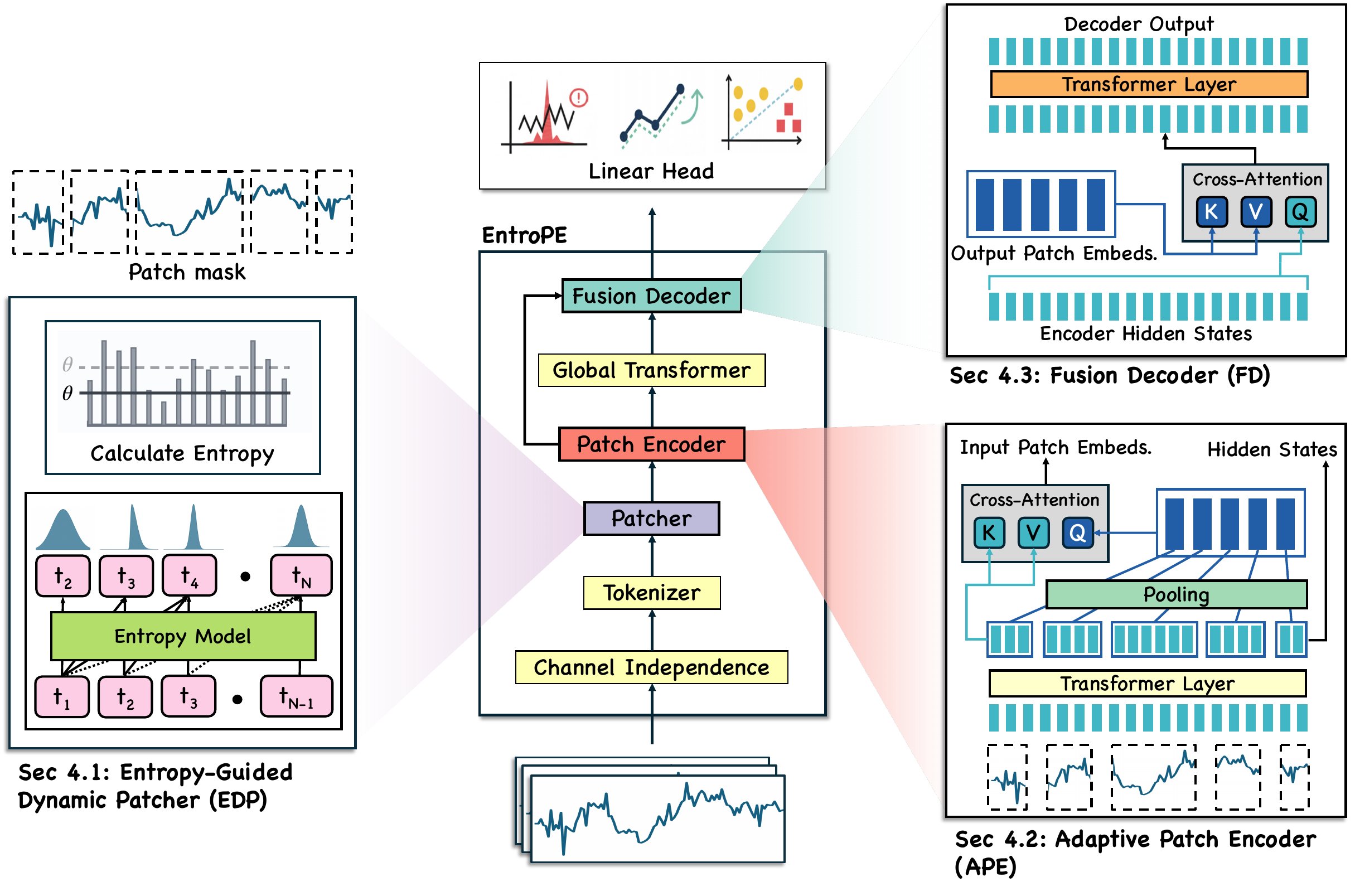}
    \caption{\texttt{EntroPE} architecture. (A) \textbf{EDP} uses a frozen causal Transformer to estimate next-token entropy and selects patch boundaries from entropy levels or increases. (B) \textbf{APE} aggregates intra-patch dependencies via pooling and cross-attention into fixed-size embeddings. (C) \textbf{FD} combines global patch context with local time-point representations.}
    \label{fig:full_arch}
\end{figure*}

Given a multivariate time series $X = [x^{(1)}, \ldots, x^{(C)}] \in \mathbb{R}^{C \times L}$ with $C$ channels and look-back length $L$, we address forecasting, classification, and anomaly detection. Following the channel-independence principle \citep{patchtst2023}, our architecture $f_\phi$ processes each channel independently to obtain $\mathbf{z}^{(c)} = f_\phi(x^{(c)})$, producing $Z = [\mathbf{z}^{(1)}, \ldots, \mathbf{z}^{(C)}] \in \mathbb{R}^{C \times d}$ that is projected to a task-specific output. The pipeline consists of entropy-guided dynamic patching, adaptive patch encoding, a global transformer over patches, and a fusion decoder, illustrated in Figure~\ref{fig:full_arch}.

\texttt{EntroPE} operates in 2 phases. \textbf{First}, we pre-train a lightweight entropy model on unlabeled data (self-supervision) to estimate next-step predictive uncertainty. \textbf{Second}, we freeze this model and use it to provide patch boundaries during task-specific training; downstream representations and output predictions remain continuous throughout.

\subsection{Entropy Model}
\label{entropy_model_pretrain}

\paragraph{Quantization.} For each univariate sequence $x = [x_1, \ldots, x_L]$, we apply a quantizer $q: \mathbb{R} \to \mathcal{V}$ with $|\mathcal{V}| = V$ to obtain tokens $\tau_t = q(x_t)$. Bin boundaries are determined from training-data statistics with $\epsilon = 0.005$ coverage:
\begin{equation}
R = \max\bigl(|Q_{\epsilon/2}(X_{\text{train}})|, |Q_{1-\epsilon/2}(X_{\text{train}})|\bigr),
\end{equation}
and $[-R, R]$ is uniformly divided into $V$ bins. Crucially, quantization is used \emph{only} to estimate the entropy signal. Unlike Chronos~\citep{ansari2024chronos}, our prediction heads operate on continuous predictions and are not constrained to discrete vocabulary decoding, sidestepping the dequantization-range limitation.

\paragraph{Architecture and Training.} A 2-layer GPT-style causal transformer~\citep{radford2019language} with 16 embedding dimension ($\approx$16K parameters) is trained for next-token prediction via cross-entropy:
\begin{equation}
\theta_{\text{ent}}^* = \arg\min_{\theta_{\text{ent}}} \mathbb{E}_{X \sim \mathcal{D}_{\text{pretrain}}} \mathcal{L}_{\text{CE}}(X; \theta_{\text{ent}}).
\end{equation}
Once converged, we freeze $\theta_{\text{ent}}^*$ and use it only to compute entropy; it is not used to generate predictions. Boundaries can be precomputed per sample. However, even computing boundaries during downstream training increased epoch duration by only about 0.1-1.0 seconds across the 6 measured datasets (Table.~\ref{tab:training_time}).

\section{EntroPE Architecture}

\subsection{Entropy-Guided Dynamic Patcher (EDP)}
\label{dynamic_patching}

For input $X \in \mathbb{R}^{C \times L}$ and channel $c$, we tokenize and compute the conditional entropy at each position:
\begin{equation}
\label{shanon_causal}
H(x_t^{(c)}) = -\sum_{v \in \mathcal{V}} p_{\theta_{\text{ent}}^*}(\tau_{t+1}^{(c)} = v \mid \tau_{\leq t}^{(c)}) \log p_{\theta_{\text{ent}}^*}(\tau_{t+1}^{(c)} = v \mid \tau_{\leq t}^{(c)}).
\end{equation}
Write $h_t^{(c)}=H(x_t^{(c)})$ for the predictive entropy at position $t$. A high value means that the frozen estimator is uncertain about the next token given the preceding tokens in channel $c$.

\paragraph{Sample-Adaptive Thresholds.}
Each forecasting configuration uses one of 2 boundary scores: the entropy level $h_t^{(c)}$ or its increase $\Delta h_t^{(c)}=h_t^{(c)}-h_{t-1}^{(c)}$. We compute the corresponding cutoff separately for each sample and channel, using the quantile level $\alpha$:
\begin{equation}
\label{eq:adaptive_thresholds}
\theta_{\mathrm{abs}}^{(c)}
  = Q_\alpha\!\left(\{h_t^{(c)}\}_{t=1}^{L}\right),
\qquad
\gamma_{\mathrm{rel}}^{(c)}
  = Q_\alpha\!\left(\{\Delta h_t^{(c)}\}_{t=2}^{L}\right).
\end{equation}
Table~\ref{tab:exp_hyperparams} gives the mode and quantile level used for each dataset and forecast horizon.

\paragraph{Boundary Placement.}
The first observation starts a patch, so $M_1^{(c)}=1$. For $1\leq t<L$, the selected rule determines whether a new patch starts at the next observation:
\begin{equation}
\label{eq:boundary_conditions}
M_{t+1}^{(c)} =
\begin{cases}
\mathbf{1}[h_t^{(c)}>\theta_{\mathrm{abs}}^{(c)}],
  & \text{entropy-level mode},\\
1, & \text{entropy-rise mode},\ t=1,\\
\mathbf{1}[\Delta h_t^{(c)}>\gamma_{\mathrm{rel}}^{(c)}],
  & \text{entropy-rise mode},\ t\geq 2.
\end{cases}
\end{equation}
Consecutive starts yield one-observation patches. When a maximum patch length is specified, we also split patches that exceed it. Let $1=b_1^{(c)}<\cdots<b_{P^{(c)}+1}^{(c)}=L+1$ denote the resulting boundaries, including the terminal endpoint. Patch $j$ contains observations $b_j^{(c)},\ldots,b_{j+1}^{(c)}-1$ and has length $\ell_j^{(c)}=b_{j+1}^{(c)}-b_j^{(c)}$.

\paragraph{Motivation for Predictive Entropy.}
With fixed-length patching, shifting the start of an input window can split the same temporal pattern at a different point. \texttt{EntroPE} uses the estimator's next-token uncertainty to guide the placement of boundaries within each window. For quantized observations, the true conditional entropy lower-bounds the expected next-token negative log-likelihood (Proposition~\ref{prop:entropy_lower_bound}). Splitting a patch preserves its length-weighted mean entropy (Proposition~\ref{prop:segmentation_optimality}). We empirically assess the practical value of the learned boundary signal through within-patch variance, boundary concentration (Section~\ref{sec:validation}) and forecasting error under alternative segmentation rules.

\subsection{Adaptive Patch Encoder (APE)}
\label{ada_patch_encoder}

The EDP produces variable-length patches whose patch count and patch length vary across samples. The APE converts these into fixed-size embeddings via pooling-initialized cross-attention, inspired by Perceiver architecture~\citep{jaegle2021perceiver} and adapted to continuous time-point embeddings (also used in BLT~\citep{pagnoni2025byte} for variable-length encoding in language modeling).

We map each quantized observation to a token embedding $\mathbf{e}_t^{(c)}=E(q(x_t^{(c)}))$. The local encoder transforms these embeddings into token states $\mathbf{h}_t^{(c)}$. For patch $j$, we initialize its representation by max-pooling the states at positions $t\in\mathcal{T}_j^{(c)}$, then refine it by cross-attention over those positions.
\begin{equation}
\mathbf{z}_j^{(c),0} = \text{MaxPool}\bigl(\{\mathbf{h}_t^{(c)}\}_{t \in \mathcal{T}_j^{(c)}}\bigr), \qquad
\mathbf{z}_j^{(c),n} = \mathbf{z}_j^{(c),n-1} + W_o \!\!\sum_{i \in \mathcal{T}_j^{(c)}} \!\!\alpha_{ji}^{(n)} W_v\bigl(\mathbf{h}_i^{(c),n-1}\bigr),
\end{equation}
with attention weights $\alpha_{ji}^{(n)}$ computed by softmax over $\mathcal{T}_j^{(c)}$. The output is patch embeddings $P^{(c)} \in \mathbb{R}^{P^{(c)} \times d_p}$ and refined token states $H^{(c)} \in \mathbb{R}^{L \times d_t}$.

For batch processing, we pad to $P_{\max} = \max_{b,c} P^{(b,c)}$ and apply attention masks $\mathcal{M}_{bc,j} = \mathbb{1}[j \leq P^{(b,c)}]$ to ignore padded slots during self-attention.

\subsection{Fusion Decoder (FD)}
\label{transformer_decoder}

A standard transformer captures inter-patch dependencies, $\tilde{P}^{(c)} = \text{Transformer}(P^{(c)}, \mathcal{M}) \in \mathbb{R}^{P^{(c)} \times d_p}$. The fusion decoder then injects this global context back into the token-level representation via cross-attention, where token states query patch representations:
\begin{equation}
\tilde{H}^{(c)} = H^{(c)} + W_o \,\text{softmax}\!\left(\frac{W_q(H^{(c)}) W_k(\tilde{P}^{(c)})^\top}{\sqrt{d_k}}+A^{(c)}\right) W_v(\tilde{P}^{(c)}).
\end{equation}
Here $A^{(c)}$ is the token-to-patch attention mask, with zero for allowed pairs and $-\infty$ otherwise. With a decoder window of one patch, each token attends to its own patch representation, which already incorporates inter-patch context from the global Transformer. The decoder further processes the fused token states with a local Transformer block. 

For forecasting, we flatten the decoder output, apply a linear projection, and reverse the instance normalization to obtain predictions on the original scale. We give the classification and anomaly-detection heads and their training objectives in Appendix~\ref{app:task_specific_heads_details}.

\section{Experiments}
\label{sec:experiments}

We evaluate \texttt{EntroPE} on long-term forecasting, classification, and anomaly detection tasks. We use the same backbone architecture across all 3 tasks; only the task-specific head changes. Section~\ref{sec:forecasting} presents the main forecasting results; Section~\ref{sec:other_tasks} demonstrates generalization to classification and anomaly detection; Section~\ref{sec:validation} empirically validates the entropy mechanism; and Section~\ref{sec:ablations} isolates and analyzes each architectural component. Unless otherwise stated, all experimental results are averaged across 3 random seeds. Full implementation settings, dataset descriptions, and extended result tables are presented in the Appendix (\ref{app:dataset_details},\ref{app:implementation},\ref{app:extended_results}).

\subsection{Long-Term Forecasting}
\label{sec:forecasting}

\textbf{Datasets.} To examine forecasting performance we conduct comprehensive evaluations on 8 widely-used long-term multivariate forecasting datasets, namely, the ETT family (ETTh1, ETTh2, ETTm1, ETTm2) that encompasses a range of scenarios in energy transfer technology at hourly and 15-minute resolution, Weather that offers insights into 21 meteorological variables over time, Electricity (ECL) featuring electricity consumption data from 321 clients, Solar-Energy that captures 10-minute power production records from 137 photovoltaic plants, and Traffic that records hourly road occupancy from 862 freeway sensors in the San Francisco Bay Area. ECL, Solar-Energy, and Traffic extend the evaluation to the high-dimensional, strongly periodic regime, with an order of magnitude more channels than the ETT family and weather. Table~\ref{tab:dataset-summary} presents detailed dataset descriptions.

\textbf{Baselines.} We benchmark against 13 state-of-the-art baselines covering 6 different categories.
(1) Transformer-based models: TimeMosaic \citep{ding2025timemosaic}, iTransformer \citep{liu2024itransformer}, PatchTST \citep{patchtst2023}, FEDformer \citep{zhou2022fedformer}; 
(2) CNN-based model: TimesNet \citep{wu2023timesnet}, TSLANet \citep{tslanet};
(3) MLP-based models: TimeMixer \citep{wang2024timemixer}, HDMixer \citep{huang2024hdmixer}, DLinear \citep{zeng2023transformers}; 
(4) Lightweight models: TimeBase \citep{huangtimebase}; 
(5) LLM-aligned and foundation models: Time-FFM \citep{liu2024time}, CALF \citep{liu2024calf}; 
(6) Filter-based model: FilterTS \citep{Wang_Liu_Duan_Wang_2025}. (Further details on baselines are discussed in Appendix \ref{app:baselineshighlights}.)

\begin{table*}[t]
\centering
\caption{Multivariate forecasting results, averaged over horizons $T \in \{96,192,336,720\}$ with fixed input length $L = 96$.
\capbest{Red} marks the best result and \capsecond{Blue} the second best. Per-horizon $T$ forecasting results are in Appendix Table~\ref{tab:full_results}.}
\label{tab:forecasting_results}

\tabstyle
\setlength{\tabcolsep}{1.35pt}

\resizebox{\textwidth}{!}{%
\begin{tabular}{l|cc|cc|cc|cc|cc|cc|cc|cc|cc}
\toprule

\multirow{2}{*}{\textbf{Model}}
& \multicolumn{2}{c}{ETTh1}
& \multicolumn{2}{c}{ETTh2}
& \multicolumn{2}{c}{ETTm1}
& \multicolumn{2}{c}{ETTm2}
& \multicolumn{2}{c}{Weather}
& \multicolumn{2}{c}{Solar}
& \multicolumn{2}{c}{Electricity}
& \multicolumn{2}{c}{Traffic}
& \multicolumn{2}{c}{\textbf{Overall}}
\\

& MSE & MAE
& MSE & MAE
& MSE & MAE
& MSE & MAE
& MSE & MAE
& MSE & MAE
& MSE & MAE
& MSE & MAE
& MSE & MAE
\\

\midrule

\textbf{\oursbold{}}
& \best{0.427}   & \second{0.431}
& \best{0.369}   & \second{0.402}
& \best{0.379}   & \second{0.394}
& 0.285          & 0.332
& \best{0.247}   & \second{0.277}
& \best{0.239}   & 0.275
& 0.193          & 0.284
& 0.498          & 0.298
& \best{0.330}   & 0.336
\\
\midrule
TimeMosaic
& \second{0.437} & \best{0.426}
& \second{0.375} & \best{0.393}
& 0.389          & \best{0.384}
& 0.283   & \best{0.319}
& 0.257          & \best{0.273}
& 0.268          & 0.269
& 0.199          & 0.276
& 0.485          & 0.290
& 0.336 & \best{0.329}
\\
CALF
& 0.443         & 0.435 
& 0.379         & 0.402 
& 0.403         & 0.397 
& 0.285         & 0.324 
& 0.257         & 0.277 
& \second{0.242}    & \best{0.243} 
& 0.192             & \second{0.271} 
& \second{0.467}    & \best{0.285} 
& \second{0.333}    & \second{0.329} 
\\
iTransformer
& 0.457          & 0.449
& 0.396          & 0.414
& 0.421          & 0.419
& 0.293          & 0.335
& 0.261          & 0.281
& 0.244 & \second{0.268}
& \best{0.182}   & \best{0.271}
& \best{0.442}   & \second{0.289}
& 0.337          & 0.341
\\
TimeMixer
& 0.483          & 0.454
& 0.397          & 0.413
& 0.389          & 0.401
& \second{0.282}   & 0.326
& \second{0.250} & 0.278
& 0.250          & 0.287
& \second{0.186} & 0.276
& 0.495          & 0.310
& 0.341          & 0.343
\\
FilterTS
& 0.446          & 0.434
& 0.381          & 0.402
& 0.389          & 0.398
& \best{0.282}   & \second{0.324}
& 0.258          & 0.282
& 0.274          & 0.299
& 0.200          & 0.284
& 0.487          & 0.326
& 0.339          & 0.344
\\
TSLANet
& 0.464          & 0.449
& 0.381          & 0.406
& \second{0.383} & 0.398
& 0.286          & 0.330
& 0.256          & 0.279
& 0.263          & 0.284
& 0.201          & 0.284
& 0.489          & 0.307
& 0.340          & 0.342
\\
TimeFFM
& 0.459          & 0.439
& 0.375          & 0.402
& 0.402          & 0.397
& 0.290          & 0.335
& 0.273          & 0.292
& 0.375          & 0.386
& 0.224          & 0.310
& 0.669          & 0.432
& 0.383          & 0.374
\\
HDMixer
& 0.452          & 0.438
& 0.383          & 0.405
& 0.398          & 0.402
& 0.284 & 0.329
& 0.256          & 0.285
& 0.313          & 0.336
& 0.199          & 0.301
& 0.543          & 0.379
& 0.353          & 0.359
\\
TimeBase
& 0.450          & 0.438
& 0.394          & 0.415
& 0.438          & 0.432
& 0.295          & 0.338
& 0.256          & 0.285
& 0.402          & 0.411
& 0.247          & 0.337
& 0.729               & 0.392
& 0.401          & 0.381
\\
FEDformer
& 0.443          & 0.460
& 0.446          & 0.457
& 0.455          & 0.460
& 0.308          & 0.353
& 0.320          & 0.371
& 0.254          & 0.363
& 0.218          & 0.333
& 0.606          & 0.374
& 0.381          & 0.396
\\
PatchTST
& 0.461          & 0.455
& 0.395          & 0.416
& 0.396          & 0.409
& 0.294          & 0.338
& 0.256          & 0.279
& 0.252          & 0.294
& 0.201          & 0.292
& 0.476 & 0.304
& 0.341          & 0.348
\\
DLinear
& 0.462          & 0.458
& 0.569          & 0.522
& 0.404          & 0.408
& 0.355          & 0.402
& 0.266          & 0.319
& 0.328          & 0.399
& 0.211          & 0.299
& 0.627          & 0.388
& 0.403          & 0.399
\\
TimesNet
& 0.477          & 0.465
& 0.414          & 0.428
& 0.404          & 0.414
& 0.298          & 0.335
& 0.262          & 0.288
& 0.345          & 0.365
& 0.251          & 0.339
& 0.652          & 0.358
& 0.388          & 0.374
\\

\bottomrule
\end{tabular}}
\end{table*}

\textbf{Results.} \texttt{EntroPE} achieves the best MSE performance on 6 out of 9 cases, including the overall average performance across all datasets (Table~\ref{tab:forecasting_results}). Compared with PatchTST, \texttt{EntroPE} reduces average MSE by approximately 3.2\% while using 90.9\% fewer parameters and 78.5\% fewer counted MACs across the 8 datasets, retaining a channel-independent design.
The other cases where \texttt{EntroPE} is not first are illustrative: on Electricity (321 channels), iTransformer and CALF dominate via cross-channel attention; on ETTm2, TimeMosaic uses larger models (\textasciitilde8M parameters versus our $\leq$ 0.8M). \texttt{EntroPE} is competitive with CALF (\textasciitilde180M) and TimeMosaic (\textasciitilde8M) at $10\text{-}225\times$ fewer parameters, indicating that gains arise from boundary placement rather than capacity. Compared to the other adaptive patching method HDMixer, \texttt{EntroPE} consistently outperforms it at matched compute. Further details on efficiency and model size are demonstrated in Figure~\ref{fig:fig_MAC_params_analysis}-~\ref{fig:fig_accuracy_vs_latency} and Table~\ref{tab:efficiency}.

\begin{wrapfigure}{r}{0.5\textwidth}
  \centering
  \vspace{-6pt}
  \captionsetup{width=0.9\linewidth}
  \includegraphics[width=0.45\textwidth]{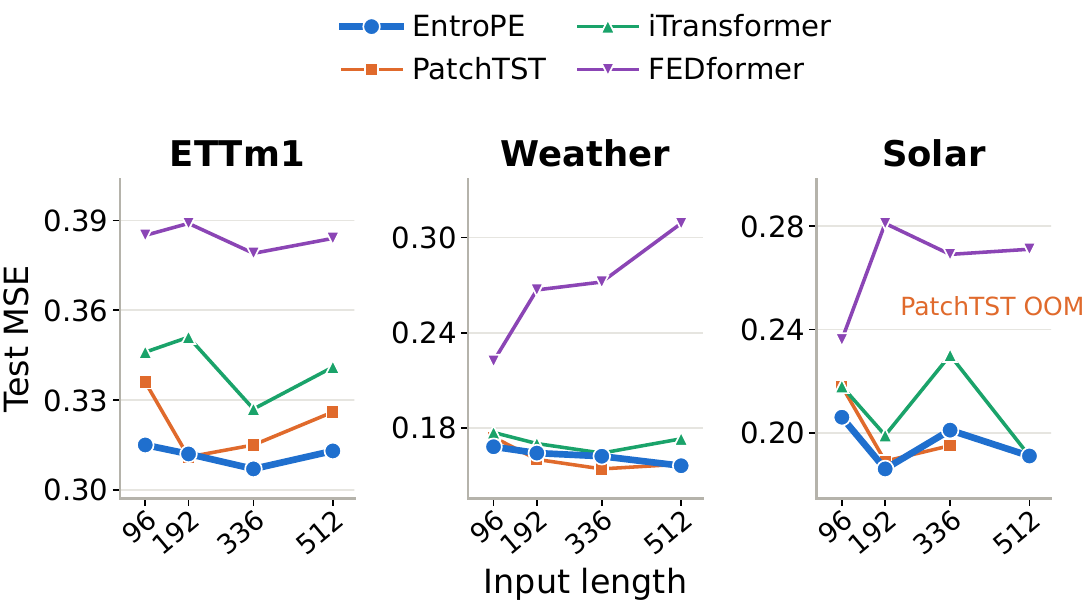}
  \caption{Test MSE for input lengths.}
    \label{fig:fig_longer_contextw}
\end{wrapfigure}
\paragraph{Longer Input Context Forecasting.} Figure~\ref{fig:fig_longer_contextw} examines forecasting performance of ETTm1, Weather and Solar as the input length increases. We additionally evaluate $L=512$ with horizons $T=96$ and $480$, and $T=720$ on Solar, comparing dynamic and static patching at small, medium, and large model sizes. Within each comparison, we fix the training budget, and the static patch length matches the average dynamic patch length. At $512\!\rightarrow\!96$, dynamic patching reduces MSE across all 5 evaluated datasets and all 3 model sizes, with reported reductions of $1.9$-$12.7\%$. At longer horizons, the gains depend on the dataset and model size, with some settings favoring static patching. Complete MSE/MAE results and relative changes are provided in Appendix~\ref{sec:rb:long_context}.

\subsection{Classification and Anomaly Detection}
\label{sec:other_tasks}

\textbf{Classification.} On 14 UEA multivariate classification benchmarks~\citep{bagnall2018uea} with the forecasting head replaced by a classification head. As demonstrated by Table~\ref{tab:classification_main}, \texttt{EntroPE} achieves $76.01\%$ average accuracy, ahead of TSLANet ($72.65\%$), TS-TCC~\citep{ijcai2021-324} ($70.98\%$), ROCKET~\citep{dempster2020rocket} ($65.18\%$), PatchTST ($68.53\%$), and GPT4TS~\citep{zhou2023one} ($66.72\%$). The dynamic patcher captures discriminative temporal patterns without architectural changes. We also include the Cricket, EigenWorms, MotorImagery, and StandWalkJump datasets from the UEA repository, which further demonstrate that \texttt{EntroPE} supports very long input contexts.

\begin{table*}[h!]
\centering
\caption{Classification accuracy (\%) on selected UEA multivariate datasets.
\capbest{Red} marks the best result and \capsecond{Blue} the second best.}
\label{tab:classification_main}
\tabstyle
\setlength{\tabcolsep}{2.0pt}   
\resizebox{0.80\textwidth}{!}{%
\begin{tabular}{l|ccccccccc}
\toprule
\multirow{1}{*}{\textbf{Dataset}}
& \textbf{\ours} & TSLANet & GPT4TS & TimesNet
& ROCKET & CrossF. & PatchTST & MLP
& TS-TCC \\
\midrule
EthanolConcentration & 29.10          & 30.42          & 25.48 & 27.73 & \best{42.58} & \second{34.98} & 28.90 & 33.46 & 32.32 \\
FaceDetection        & \best{69.10}   & 66.77          & 65.58 & 67.47 & 64.70 & 66.17 & \second{68.96} & 67.42 & 63.05 \\
Handwriting          & \second{57.40} & \best{57.88}   & 34.56 & 26.18 & 48.47 & 26.24 & 26.00 & 22.47 & 47.76 \\
Heartbeat            & \best{80.00}   & \second{77.56} & 36.59 & 74.48 & 69.76 & 76.59 & 76.59 & 73.17 & 77.07 \\
JapaneseVowels       & \best{99.20}   & \second{99.19} & 98.11 & 97.83 & 95.68 & 98.92 & 98.65 & 97.84 & 97.30 \\
PEMS-SF              & \best{88.60}   & 83.82          & 87.28 & 88.13 & 75.10 & 82.08 & \second{88.44} & 82.08 & 86.71 \\
SelfRegulationSCP1   & \best{92.90}   & 91.81          & 91.47 & 77.43 & 84.64 & \second{92.49} & 89.76 & 88.40 & 91.13 \\
SelfRegulationSCP2   & \best{63.10}   & \second{61.67} & 51.67 & 52.84 & 54.44 & 53.33 & 54.44 & 51.67 & 53.89 \\
SpokenArabicDigits   & \second{99.80} & \best{99.91}   & 99.36 & 98.36 & 99.20 & 96.41 & 99.68 & 96.68 & 99.77 \\
UWaveGestureLibrary  & \best{94.40}   & \second{91.25} & 84.38 & 83.13 & 76.03 & 81.56 & 80.00 & 81.88 & 86.25 \\
Cricket              & 83.33          & \second{98.61} & 94.44 & 91.67 & \best{100.00} & 91.67 & 95.83 & 75.00 & 94.44 \\
EigenWorms           & \best{74.81}   & \second{72.52} & 56.49 & 53.44 & 9.92 & 52.67 & 46.56 & 54.20 & 47.33 \\
MotorImagery         & 59.00          & 59.00 & \best{62.00} & \second{61.00} & 52.00 & 50.00 & 59.00 & 47.00 & 50.00 \\
StandWalkJump        & \best{73.33}   & 26.67 & 46.67 & 60.00 & 40.00 & 46.67 & 46.67 & 53.33 & \second{66.67} \\
\midrule
\textbf{Average}     & \best{76.01}   & \second{72.65} & 66.72 & 68.55 & 65.18 & 67.84 & 68.53 & 66.04 & 70.98 \\
\bottomrule
\end{tabular}}
\end{table*}
 
\textbf{Anomaly Detection.} As demonstrated by Table~\ref{tab:anomaly_main}, on 5 standard anomaly detection benchmarks (SMD, MSL, SMAP, SWaT, PSM), \texttt{EntroPE} matches or exceeds baselines on F1 score. We also include ETSformer~\citep{woo2022etsformer} and LightTS~\citep{lights} as baselines. These datasets have a 100-length input context window. Therefore, we train the entropy model with a block size of 100 for anomaly detection.

\begin{table}[h!]
\centering
\caption{Anomaly detection F1 score (\%) on 5 benchmark datasets.
\capbest{Red} marks the best result and \capsecond{Blue} the second best.}
\label{tab:anomaly_main}
\tabstyle
\setlength{\tabcolsep}{2pt}   
\resizebox{0.83\textwidth}{!}{%
\begin{tabular}{l|ccccccccc}
\toprule
\textbf{Dataset} & \textbf{\ours} & TSLANet & GPT4TS
& TimesNet & PatchTST & ETSformer
& FEDformer & LightTS  & DLinear \\
\midrule
SMD  & \second{87.10} & \best{87.91}   & 86.89          & 84.61 & 84.62 & 83.13          & 85.08 & 82.53 & 77.10 \\
MSL  & \best{85.30}   & 83.32          & 82.45          & 81.84 & 78.70 & \second{85.03} & 78.57 & 78.95 & 84.88 \\
SMAP & \best{77.93} & \second{75.96} & 72.88 & 69.39 & 68.82 & 69.50 & 70.76 & 69.21 & 69.26 \\

SWaT & \best{94.30}   & 92.80          & \second{94.23} & 93.02 & 85.72 & 84.91          & 93.19 & 93.33 & 87.52 \\
PSM  & \second{97.70} & \best{97.73}   & 97.13          & 97.34 & 96.08 & 91.76          & 97.23 & 97.15 & 93.55 \\
\midrule
\textbf{Average} & \best{88.47} & \second{87.54} & 86.72 & 85.24 & 82.79 & 82.87 & 84.97 & 84.23 & 82.46 \\
\bottomrule
\end{tabular}}
\end{table}

\subsection{Empirical Validation of Entropy-Guided Segmentation}
\label{sec:validation}
\begin{wraptable}{r}{0.52\textwidth}
\vspace{-6pt}
\centering
\caption{Patch-level diagnostics on 500 test samples.}
\label{tab:validation}
\tabstyle
\setlength{\tabcolsep}{2.5pt}
\renewcommand{\arraystretch}{1.1}  
\begin{adjustbox}{max width=\linewidth}
\begin{tabular}{l|cc|cc}
\toprule
\multirow{2}{*}{\textbf{Dataset}} & \multicolumn{2}{c}{\textbf{Variance ratio} $\downarrow$}
& \multirow{2}{*}{\textbf{Gain}} & \multirow{2}{*}{\textbf{Boundary conc.}} \\
& Dynamic & Static & & \\
\midrule
ETTh1       & 0.195 $\pm$ 0.10 & 0.516 $\pm$ 0.20 & 2.6$\times$ & 0.974 $\pm$ 0.02 \\
ETTh2       & 0.223 $\pm$ 0.22 & 0.481 $\pm$ 0.26 & 2.2$\times$ & 0.977 $\pm$ 0.02 \\
ETTm1       & 0.066 $\pm$ 0.06 & 0.208 $\pm$ 0.13 & 3.1$\times$ & 0.971 $\pm$ 0.02 \\
ETTm2       & 0.111 $\pm$ 0.11 & 0.307 $\pm$ 0.22 & 2.8$\times$ & 0.978 $\pm$ 0.02 \\
Electricity & 0.172 $\pm$ 0.09 & 0.484 $\pm$ 0.26 & 2.8$\times$ & 0.957 $\pm$ 0.01 \\
Weather     & 0.052 $\pm$ 0.08 & 0.175 $\pm$ 0.21 & 3.4$\times$ & 0.977 $\pm$ 0.02 \\
\midrule
\textbf{Mean} & \textbf{0.137} & \textbf{0.362}
             & \textbf{2.8}\(\times\) & \textbf{0.972} \\
\bottomrule
\end{tabular}
\end{adjustbox}
\end{wraptable}
We test whether entropy-guided boundaries produce the patch-level properties that motivate the design. On 500 test samples per dataset, we measure 2 quantities. (1) The \emph{intra-patch variance ratio} is the average within-patch variance divided by the total sequence variance; lower values indicate predictively coherent patch interiors. (2) The \emph{boundary entropy concentration} is the fraction of detected boundaries that coincide with the top-$\alpha$ highest-entropy positions. Across all 6 datasets, dynamic patches are $2.8\times$ more internally homogeneous than fixed patches, and $97.2\%$ of boundaries fall at high-entropy positions (Table~\ref{tab:validation}). The mechanism behaves as the design intends: predictable regimes are grouped into patches, while high-uncertainty transitions become boundaries.

\subsection{Ablation Studies}
\label{sec:ablations}

\begin{figure}[h!]
\centering
    \includegraphics[width=1\textwidth]{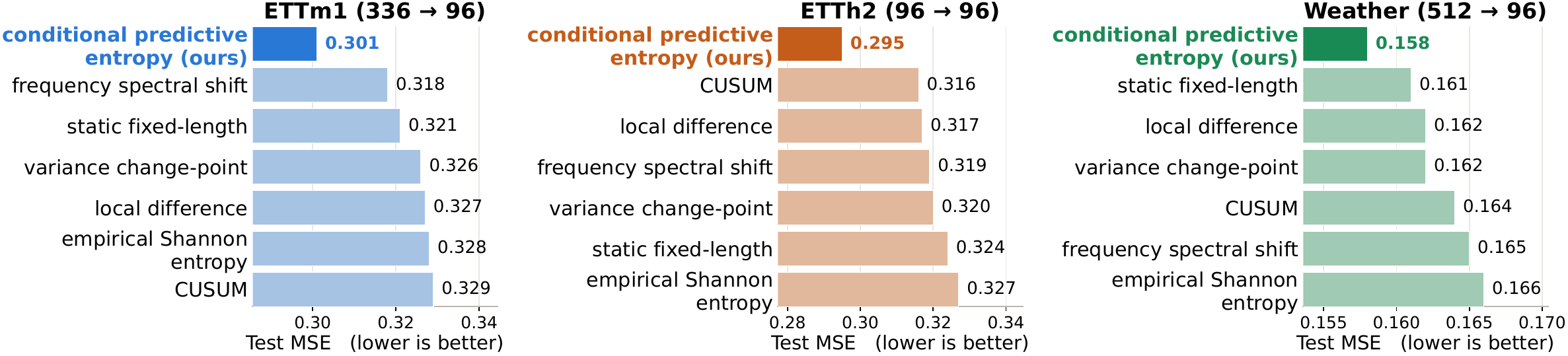}
    \caption{Boundary-signal ablation on ETTm1, ETTh2 and Weather different input lengths.}
    \label{fig:boundary_ablation}
    \vspace{-6pt}
\end{figure}

\paragraph{Boundary Signal Analysis.} A core question is whether \emph{conditional predictive entropy} is meaningfully different from simpler change-point heuristics. We hold the architecture fixed (all variants share the same APE, global transformer, and FD) and replace only the boundary signal, comparing 6 alternatives on ETTm1, ETTh2, and Weather with 3 different input context lengths $L \in \{96,336,512\}$. Figure~\ref{fig:boundary_ablation} shows that conditional predictive entropy outperforms variance change-point, frequency spectral shift, CUSUM, local difference, EAPformer-style empirical Shannon entropy~\citep{ling2025eapformer}, and static boundaries under matched conditions.

\paragraph{Architectural Components Analysis.} We progressively remove the EDP, APE, and FD on 4 datasets at prediction horizons $T \in \{336, 720\}$. The full architecture wins 7 of 8 cases (Table~\ref{tab:comprehensive_ablation}); removing the EDP (replacing dynamic patches with static ones) confirms that boundary placement contributes more than encoder design or decoder fusion. This ablation experiment used a single seed MSE on each run.

\begin{table}[h!]
\centering
\caption{Component ablation (MSE $\downarrow$).
\capbest{Red} marks the best result and \capsecond{Blue} the second best.}
\label{tab:comprehensive_ablation}
\tabstyle
\setlength{\tabcolsep}{3pt}
\resizebox{0.65\textwidth}{!}{%
\begin{tabular}{ll|c|ccc|c|c}
\toprule
\multirow{2}{*}{\textbf{Dataset}} & \multirow{2}{*}{\textbf{$T$}}
& \textbf{Full} & \multicolumn{3}{c}{$-$\textbf{EDP}}
& $-$\textbf{EDP}$-$\textbf{APE} & $-$\textbf{EDP}$-$\textbf{APE}$-$\textbf{FD} \\
& & Dyn. & Stat(1) & Stat(8) & Stat(16) & Pool+FD & Pool+Flat \\
\midrule
\multirow{2}{*}{ETTh1}   & 336 & \second{0.429} & \best{0.425}   & 0.441 & 0.444 & 0.438 & 0.519 \\
                         & 720 & \best{0.439}   & \second{0.460} & 0.469 & 0.477 & 0.461 & 0.527 \\
\addlinespace[2pt]
\multirow{2}{*}{ETTh2}   & 336 & \best{0.355}   & \second{0.428} & 0.435 & 0.439 & 0.439 & 0.462 \\
                         & 720 & \best{0.397}   & \second{0.448} & 0.470 & 0.463 & 0.466 & 0.456 \\
\addlinespace[2pt]
\multirow{2}{*}{ETTm1}   & 336 & \best{0.393}   & \second{0.401} & 0.409 & 0.415 & 0.402 & 0.427 \\
                         & 720 & \best{0.445}   & \second{0.451} & 0.459 & 0.452 & 0.457 & 0.483 \\
\addlinespace[2pt]
\multirow{2}{*}{Weather} & 336 & \best{0.258}   & \second{0.261} & 0.265 & 0.270 & 0.262 & 0.262 \\
                         & 720 & \best{0.341}   & \second{0.343} & 0.353 & 0.359 & 0.347 & 0.346 \\
\bottomrule
\end{tabular}}
\end{table}

\begin{figure}[h!]
\centering
\includegraphics[width=1\textwidth]{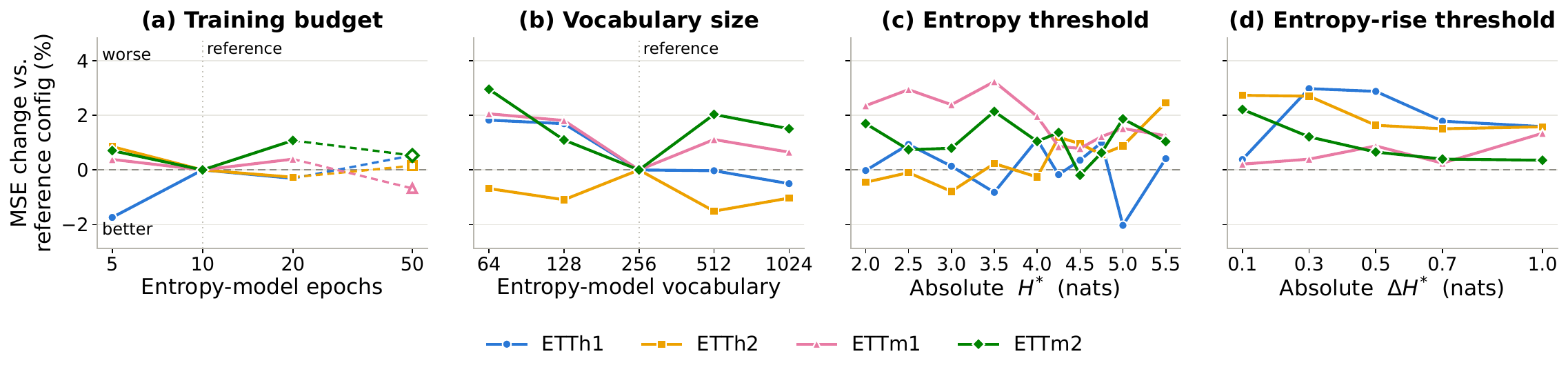}
\caption{Sensitivity of \texttt{EntroPE} to the entropy model and its thresholding rules.}
\label{fig:fig_entropy_ablation}
\end{figure}

\paragraph{Entropy Model Training and Threshold Analysis.} 
In Figure~\ref{fig:fig_entropy_ablation}, each panel varies one control on the ETT-family datasets at input length $L = 96$; the y-axis is the change in test MSE against the reference configuration, averaged over $T \in {96, 336}$ (negative is better). The reference is a vocabulary-256 entropy model trained for 10 epochs with the per-dataset percentile threshold, so the curves pass through zero at 10 epochs in (a) and at vocabulary 256 in (b). Extending training to 20 or 50 epochs yields no consistent forecasting gain. Most vocabulary settings change MSE by at most about 2\%, with the largest increase near 3\% on ETTm2 at $V=64$. The effect of a fixed absolute threshold depends on the dataset. For example, a threshold of 5.0 nats reduces MSE by about 2\% on ETTh1 but increases it on the other 3 datasets. For the entropy-rise rule, all tested fixed thresholds yield higher MSE than the percentile reference, although the gap stays below 2\% on all 4 datasets at 0.7 and 1.0 nats.

\paragraph{Efficiency Analysis.} 
\begin{figure}[tbp]
    \centering
    \begin{minipage}[t]{0.49\linewidth}
        \vspace{0pt}
        \centering
        \includegraphics[width=\linewidth]{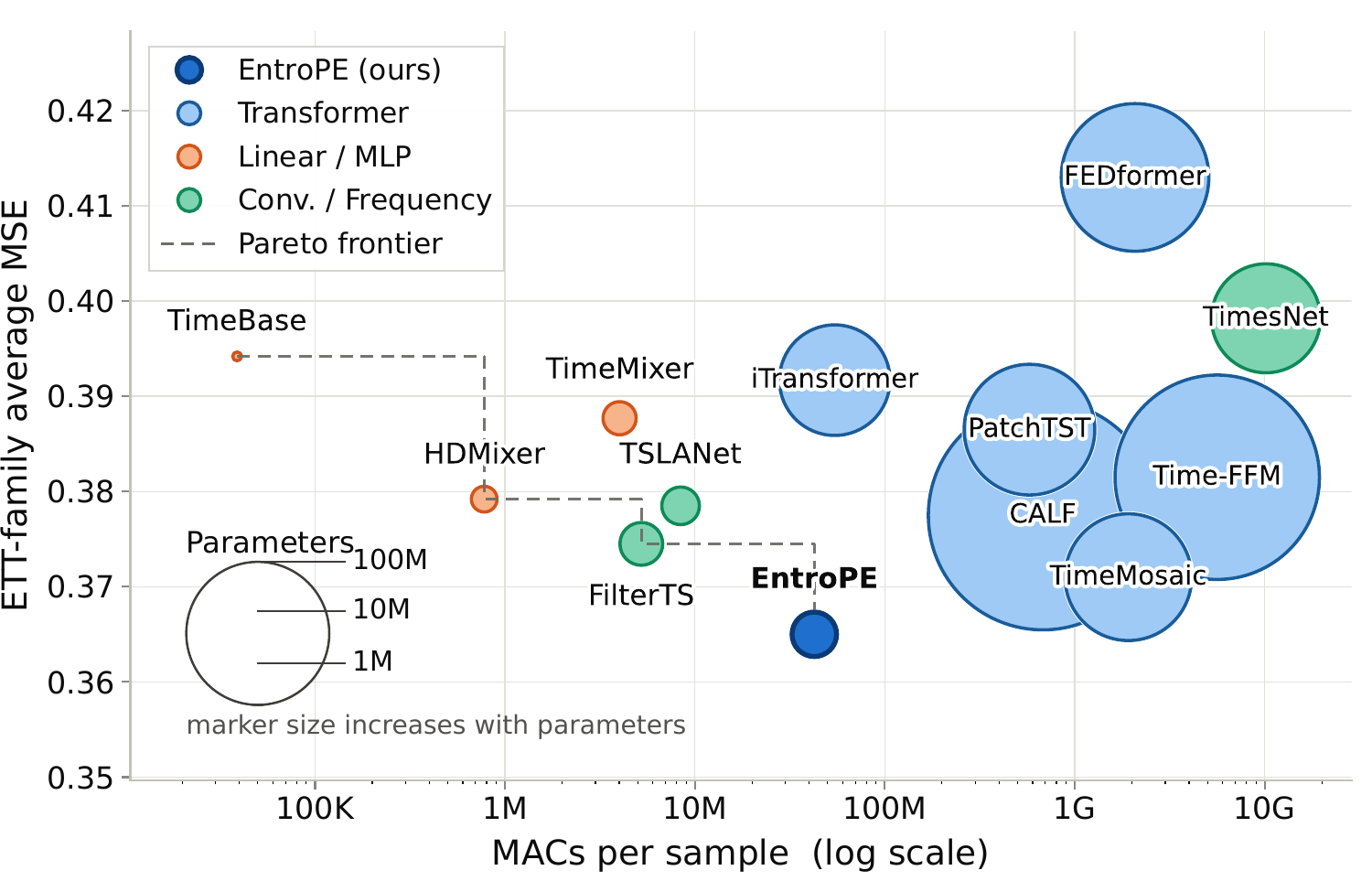}
        \caption{ETT-family average MSE versus counted MACs per
        sample. Bubble size indicates parameter count.}
        \label{fig:fig_MAC_params_analysis}
    \end{minipage}\hfill%
    \begin{minipage}[t]{0.49\linewidth}
        \vspace{0pt}
        \centering
        \includegraphics[width=\linewidth]{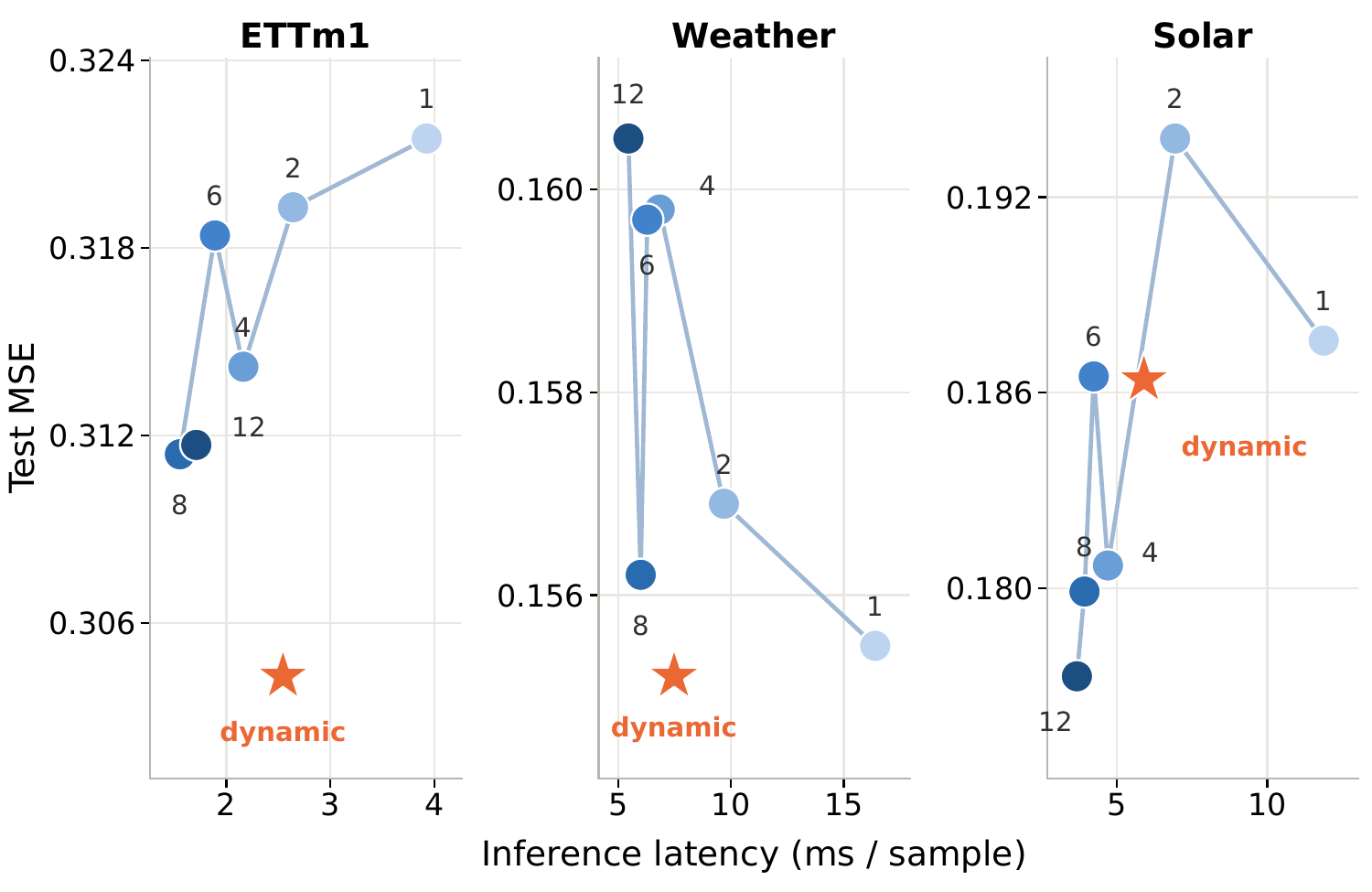}
        \caption{Test MSE versus inference latency per sample
        for dynamic and fixed-length patching at $T=96$.}
        \label{fig:fig_accuracy_vs_latency}
    \end{minipage}
\end{figure}
The quantile threshold $\alpha$ acts as a computational control knob: lower values yield finer segmentation, higher values produce coarser patches. Figure~\ref{fig:fig_MAC_params_analysis} compares forecasting error and computational cost across the 4 ETT datasets. \texttt{EntroPE} achieves the lowest ETT-family average MSE among the plotted models at roughly 42 million counted MACs per sample. Models such as HDMixer and FilterTS use fewer MACs but have higher average MSE, placing \texttt{EntroPE} at the low-error end of the observed Pareto frontier.

Figure~\ref{fig:fig_accuracy_vs_latency} shows at forecast horizon $T = 96$ and lookbacks are ETTm1 $L = 336$; Weather $L = 512$; Solar $L = 192$. Numbered points are static patching at fixed patch length $P \in {1, 2, 4, 6, 8, 12}$; the highlighted point is \texttt{EntroPE}'s entropy-based dynamic patching ($\text{threshold} = 0.95$). 
Timing covers the full forward pass, including entropy estimation and boundary generation; the static variants bypass the entropy model entirely, so the comparison is conservative with respect to dynamic patching. On ETTm1 and Weather, dynamic patching achieves lower test MSE than all 6 tested fixed patch lengths and runs faster than the finest fixed segmentation ($P=1$). On Solar, the fixed patches of length $P=12$ achieve both lower MSE and lower latency than dynamic patching.
 
\begin{figure}[h!]
\centering
    \includegraphics[width=1\textwidth]{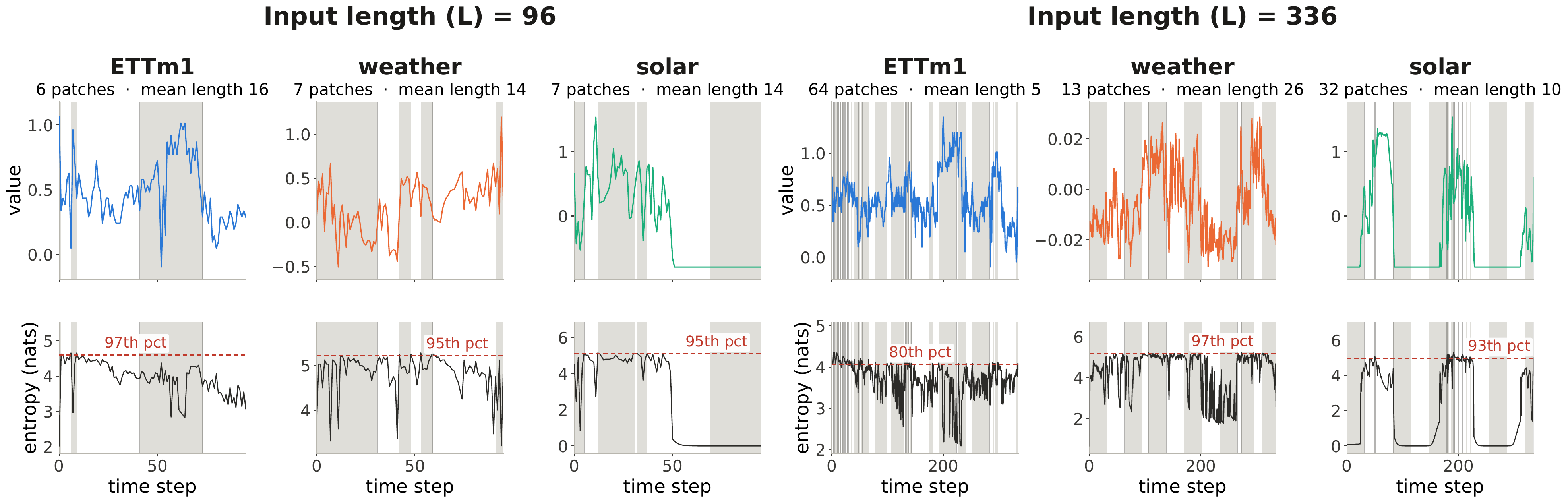}
    \caption{Entropy-guided patch placement on example windows from ETTm1, Weather, and Solar.
    }
    \label{fig:dynamic_patch_visual}
\end{figure}

\paragraph{Qualitative Analysis.} Figure~\ref{fig:dynamic_patch_visual} illustrates how patch placement varies across and within input windows. The upper panels show the input signal and the lower panels show estimated next-token entropy in nats. Alternating shaded and unshaded regions mark successive patches. The dashed red line indicates the absolute entropy threshold. At $L=96$, the ETTm1 and Weather examples contain 6 and 7 patches, with mean lengths of 16 and 14 time steps, respectively, while the Solar example contains 7 patches with a mean length of 14. The longer Solar example ($L=336$) shows how patch spacing changes within a single window. Several boundaries cluster around the burst of activity near time step 200, where estimated predictive entropy is high. Boundaries are sparse across the long, nearly flat intervals where entropy is low. This is a strong example of how computation is allocated based on flat (low information density) regions.

\section{Limitations}
\label{sec:limitations}
 \texttt{EntroPE}'s benefits depend on the dataset, forecast horizon, and model size. Improvements are small and mixed on M4-Monthly ($36\!\rightarrow\!18$), while static patching achieves lower MSE on Solar for the small model at horizons 480 and 720 and the large model at horizon 720 (Tables~\ref{tab:rb:long_context}-\ref{tab:rb:relative_changes}). Boundary placement is also sensitive to noise: on ETTh1, Jaccard similarity to clean-signal boundaries decreases from $0.63$ at 40 dB SNR to $0.28$ at 2 dB, despite a stable boundary count (Table~\ref{tab:noise_boundary_stability}). The dataset-specific entropy estimator may require retraining under domain shifts, and channel-independent segmentation can miss coordinated multivariate transitions. Furthermore, recurring boundaries on strongly periodic datasets suggest that simpler period-based segmentation may provide comparable benefits. Future work includes noise-robust and transferable entropy estimators, cross-channel boundary coordination, and mechanisms that adaptively select between dynamic, fixed, and period-based segmentation according to signal characteristics and computational constraints.

\section{Conclusion}
We introduced \texttt{EntroPE}, a dynamic patch encoding framework that uses entropy-guided boundary detection to identify natural temporal transitions in time series. By placing patch boundaries at information-rich transitions rather than fixed intervals, \texttt{EntroPE} preserves temporal coherence while enabling efficient variable-length processing. This information-theoretic approach improves performance across downstream tasks with practical computational benefits. Beyond empirical gains, \texttt{EntroPE} demonstrates the value of respecting intrinsic temporal structure, opening new directions in adaptive context-aware sequence modeling.

\section*{Acknowledgments}
We thank the anonymous reviewers for their helpful feedback. Sachith Abeywickrama acknowledges support from the Agency for Science, Technology and Research (A*STAR), Singapore, through the Singapore International Graduate Award (SINGA). This research is supported by the Ministry of Education, Singapore, under the MOE Tier 1 grant (SKI 2021\_08\_03), and by the Kwan Im Thong Hood Cho Temple Chair Professorship. This research is also supported by the KU Research Center for Advanced Intelligent Systems (KU-AIS) at Khalifa University of Science and Technology.

\bibliographystyle{plainnat}
\bibliography{neurips_2026}


\clearpage
\appendix
\section*{Appendix}

Sections~\ref{app:dataset_details} and~\ref{app:implementation} describe the benchmarks and implementation. Section~\ref{app:extended_results} presents the forecasting configurations and results. Sections~\ref{app:seed_analyses} and~\ref{app:efficiency_details} report seed sensitivity and computational cost. Section~\ref{sec:rb:patch_statistics} examines how the segmentation behaves across samples, thresholds, noise levels, and datasets.

\section{Datasets and baselines}
\label{app:dataset_details}

\subsection{Forecasting Datasets}
\label{app:forecast_datasets}

Table~\ref{tab:dataset-summary} summarizes the 8 datasets in the main forecasting comparison. ETTh1 and ETTh2 record transformer temperature and load hourly, while ETTm1 and ETTm2 record them every 15 minutes. Weather contains 21 meteorological variables. Electricity (ECL) records hourly consumption for 321 clients. Solar records generation at 137 photovoltaic plants every 10 minutes, and Traffic records hourly road occupancy at 862 freeway sensors. The additional monthly forecasting experiment is described in Section~\ref{sec:rb:long_context}.

\begin{table*}[!htbp]
\centering
\small
\caption{Forecasting datasets in the main comparison. Split sizes are listed in training, validation, and test order.}
\label{tab:dataset-summary}
\renewcommand{\arraystretch}{1.2}
\resizebox{\textwidth}{!}{%
\begin{tabular}{lcccc}
\toprule[1pt]
\textbf{Dataset} & \textbf{Dim} & \textbf{Reported split sizes (train, val., test)} & \textbf{Frequency} & \textbf{Domain} \\
\midrule
ETTh1            & 7    & (8545, 2881, 2881)     & Hourly  & Energy Infrastructure \\
ETTh2            & 7    & (8545, 2881, 2881)     & Hourly  & Energy Infrastructure \\
ETTm1            & 7    & (34465, 11521, 11521)  & 15 min   & Energy Infrastructure \\
ETTm2            & 7    & (34465, 11521, 11521)  & 15 min   & Energy Infrastructure \\
Weather          & 21   & (36792, 5271, 10540)   & 10 min   & Meteorological Monitoring \\
Electricity (ECL)& 321  & (18317, 2633, 5261)    & Hourly  & Electricity Consumption \\
Solar            & 137  & (36697, 5257, 10513)   & 10 min   & Solar Power Generation \\
Traffic          & 862  & (12185, 1757, 3509)    & Hourly  & Road Occupancy \\
\bottomrule[1pt]
\end{tabular}
}
\end{table*}
\subsection{Classification Datasets}
\label{app:classification_datasets}

We evaluate classification on 14 multivariate datasets from the UEA archive, following TSLANet~\citep{tslanet}. The collection includes Cricket, EigenWorms, MotorImagery, and StandWalkJump. Dataset descriptions are available at \url{https://www.timeseriesclassification.com/}.

\subsection{Anomaly Detection Datasets}
\label{app:anomaly_datasets}

We follow the anomaly detection protocol of TSLANet~\citep{tslanet} and use input windows of 100 observations (Table~\ref{tab:anomaly-summary}). SMD and PSM contain server-monitoring data. MSL and SMAP contain spacecraft telemetry, and SWaT records sensor measurements from a water treatment testbed.

\begin{table*}[!htbp]
\centering
\small
\caption{Anomaly detection datasets, reported partition sizes, and input window lengths. Partition sizes are listed in training, validation, and test order.}
\label{tab:anomaly-summary}
\renewcommand{\arraystretch}{1.2}
\resizebox{0.7\textwidth}{!}{%
\begin{tabular}{lcccc}
\toprule[1pt]
\textbf{Dataset} & \textbf{Dim} & \textbf{Reported split sizes (train, val., test)} & \textbf{Length} & \textbf{Domain} \\
\midrule
SMD            & 38    & (566724, 141681, 708420)       & 100   & Server Machine \\
MSL            & 55    & (44653, 11664, 73729)          & 100   & Spacecraft \\
SMAP           & 25    & (108146, 27037, 427617)        & 100   & Spacecraft \\
SWaT           & 51    & (396000, 99000, 449919)        & 100   & Infrastructure \\
PSM            & 25    & (105984, 26497, 87841)         & 100   & Server Machine \\
\bottomrule[1pt]
\end{tabular}
}
\end{table*}

\subsection{Forecasting Baselines}
\label{app:baselineshighlights}

Table~\ref{tab:baselineshighlights} lists the 13 baselines in the main forecasting comparison and the feature most relevant to each model.

\begin{table}[!htbp]
\centering
\caption{Forecasting baselines grouped by architecture.}
\label{tab:baselineshighlights}
\scriptsize
\setlength{\tabcolsep}{3pt}
\renewcommand{\arraystretch}{1.0}
\begin{tabular}{@{}p{0.17\linewidth}p{0.24\linewidth}p{0.51\linewidth}@{}}
\toprule[1pt]
\textbf{Model} & \textbf{Category} & \textbf{Key Features / Innovations} \\
\midrule
TimeMosaic \citep{ding2025timemosaic} & Transformer & patch granularity based
on temporal heterogeneity and motif reuse. \\
iTransformer \citep{liu2024itransformer} & Transformer & Inverted attention across feature and time dimensions. \\
PatchTST \citep{patchtst2023} & Transformer & Patch-based embedding with channel-independence. \\
FEDformer \citep{zhou2022fedformer} & Transformer (freq.-decomp.) & Fourier/Wavelet enhanced decomposition. \\
\midrule
TimesNet \citep{wu2023timesnet} & CNN & Temporal 2D variation modeling with convolutional layers. \\
TSLANet \citep{tslanet} & CNN + Lightweight & Lightweight and pretraining module for enhanced feature representation \\
\midrule
TimeMixer \citep{wang2024timemixer} & MLP & Multiscale mixing, separation of trend and seasonal signals. \\
HDMixer \citep{huang2024hdmixer} & MLP & Hierarchical dependency with extendable patches. \\
DLinear \citep{zeng2023transformers} & Linear/MLP & Channel-independent trend/seasonal decomposition. \\
\midrule
Time-FFM \citep{liu2024time} & Foundation + Language & Pre-trained foundation model for time series. \\
TimeBase \citep{huangtimebase} & Lightweight & Resource-efficient, practical forecasting baseline. \\
\midrule
CALF \citep{liu2024calf} & LLM-aligned / Cross-modal & Cross-modal match, feature regularization, output consistency. \\
\midrule
FilterTS \citep{Wang_Liu_Duan_Wang_2025} & Filter-based & Static global + dynamic cross-variable frequency filtering. \\
\bottomrule
\end{tabular}
\end{table}

\FloatBarrier

\section{Entropy Estimation, Patching, and Task Objectives}
\label{app:implementation}

\subsection{Input Normalization and Tokenization}
\label{chronos_quantization}

Following Chronos~\citep{ansari2024chronos}, we represent normalized observations by discrete tokens. We compute the feature means and standard deviations for normalization on the training split. Let $X_{\mathrm{train}}$ denote the normalized training values used to estimate the range of the tokenizer. With $\epsilon=0.005$, we set
\[
 R=\max\!\left(\left|Q_{\epsilon/2}(X_{\mathrm{train}})\right|,
                  \left|Q_{1-\epsilon/2}(X_{\mathrm{train}})\right|\right).
\]
We use uniform bins over $[-R,R]$ and assign out-of-range values to the nearest end bin. For ordered bin edges $b_1,\ldots,b_{B-1}$, writing $b_0=-\infty$ and $b_B=+\infty$ gives
\begin{equation}
 q(x)=j \quad\text{if } b_{j-1}\leq x<b_j,
 \qquad j\in\{1,\ldots,B\}.
 \label{eq:quantization}
\end{equation}
If $c_j$ denotes a bin center, its representative value is $d(j)=c_j$. In the forecasting model, the quantized tokens supply inputs to the entropy estimator and to the patch encoder. The forecasting head produces continuous values from the learned representations

\subsection{Next-Token Training and Predictive Entropy}
\label{app:entropy_training}

For each dataset, we train a causal model on sequences of quantized observations $\tau_{1:L}$. Its next-token cross-entropy loss on one sequence is
\begin{equation}
 \mathcal L_{\mathrm{CE}}=-\sum_{t=1}^{L-1}\log p_\theta(\tau_{t+1}\mid\tau_{\leq t}).
 \label{eq:cross_entropy}
\end{equation}
We use AdamW with a learning rate of $10^{-3}$, weight decay of 0.05, gradient clipping at 1.0, and a batch size of 128. Training runs for up to 50 epochs, with early stopping after five epochs without improvement in validation loss. Table~\ref{tab:entropy_model} gives the forecasting estimator configuration.

For a discrete distribution $p$, Shannon entropy is $-\sum_v p(v)\log p(v)$. After training, we freeze the model and compute per-position predictive entropy in nats,
\[
 H_t=-\sum_{v\in\mathcal V}p_\theta(v\mid\tau_{\leq t})\log p_\theta(v\mid\tau_{\leq t}).
\]
A large $H_t$ means that the model assigns probability across several possible next tokens.

\begin{table}[!htbp]
\centering
\caption{Configuration of the entropy estimator used in forecasting.}
\label{tab:entropy_model}
\begin{tabular}{lc}
\toprule[1pt]
\textbf{Hyperparameter} & \textbf{Value} \\
\midrule
Number of layers ($n\_layer$)   & 2 \\
Number of heads ($n\_head$)     & 4 \\
Embedding dimension ($n\_embd$) & 16 \\
Dropout                         & 0.1 \\
Bias                            & False \\
Vocabulary size ($vocab\_size$) & 256 \\
Block size ($block\_size$)      & 96 \\
\bottomrule
\end{tabular}
\end{table}

\subsection{Boundary Rules and Inputs Longer Than One Block}
\label{sec:rb:entropy_blocks}

The setting in Table~\ref{tab:exp_hyperparams} determines which score we threshold. With \emph{Mono.} set to 0, we compare $H_t$ with the $\alpha$-quantile of entropy values in the sample. With \emph{Mono.} set to 1, we compare the increase $\Delta H_t=H_t-H_{t-1}$ with
\begin{equation}
 \gamma_{\mathrm{rel}}=Q_\alpha\!\left(\{\Delta H_t\}_{t=2}^{L}\right).
 \label{eq:adaptive_relative_threshold_appendi}
\end{equation}
A score above the selected threshold starts a new patch after that position. The quantile $\alpha$ is dimensionless. The resulting cutoff is measured in nats.

The same frozen estimator can score inputs of different lengths. For inputs longer than its 96-observation block, it processes non-overlapping blocks. In the released forecasting implementation, each sample and channel is right-padded independently to a multiple of the block size, and scores at padded positions are discarded. The estimator is reused across forecasting lookbacks and classification sequence lengths.

\subsection{Task Heads and Training Losses}
\label{app:task_specific_heads_details}

Let $\tilde H^{(c)}$ denote the fused representation for channel $c$. We use a task-specific linear head and train the forecasting, classification, and reconstruction models with the objectives below.

\paragraph{Forecasting.} For each channel, the head predicts $T$ future values.
\[
 \hat Y^{(c)}=g_{\mathrm{fc}}(\mathrm{Flatten}(\tilde H^{(c)}))\in\mathbb R^T,\qquad
 \mathcal L_{\mathrm{forecast}}
   =\frac{1}{CT}\sum_{c=1}^{C}\sum_{t=1}^{T}(Y_{c,t}-\hat Y_{c,t})^2.
\]

\paragraph{Classification.} The head projects the concatenated channel representations to $K$ class logits,
\[
 z=g_{\mathrm{cls}}\!\left(\mathrm{Flatten}([\tilde H^{(1)},\ldots,\tilde H^{(C)}])\right)
   \in\mathbb R^K,\qquad
 \mathcal L_{\mathrm{class}}=-\log\frac{\exp(z_y)}{\sum_{k=1}^{K}\exp(z_k)},
\]
where $y$ is the observed class.

\paragraph{Anomaly Detection.} The reconstruction head predicts the $L$ input values of each channel.
\[
 \hat X^{(c)}=g_{\mathrm{rec}}(\mathrm{Flatten}(\tilde H^{(c)}))\in\mathbb R^L,\qquad
 \mathcal L_{\mathrm{anomaly}}
   =\frac{1}{CL}\sum_{c=1}^{C}\sum_{t=1}^{L}(X_{c,t}-\hat X_{c,t})^2.
\]

\paragraph{Downstream Training.} We freeze the entropy estimator $\theta^*_{\mathrm{ent}}$ and optimize the token embeddings, adaptive patch encoder, global Transformer, fusion decoder, and task head. For these parameters $\Phi$, the training objective is
\[
 \Phi^*=\arg\min_\Phi\,
 \mathbb E_{(X,Y)\sim\mathcal D_{\mathrm{task}}}
 [\mathcal L_{\mathrm{task}}(X,Y;\Phi\mid\theta^*_{\mathrm{ent}})].
\]

\subsection{Information-Theoretic Motivation}
\label{sec:theory}

\begin{proposition}[Conditional entropy and next-token log loss]
\label{prop:entropy_lower_bound}
For a fixed history $h=\tau_{\leq t}$, let $p_\star(\cdot\mid h)$ denote the true next-token distribution. The expected negative log-likelihood of any predictive distribution $q(\cdot\mid h)$ is at least $H(p_\star(\cdot\mid h))$, with equality exactly when $q(\cdot\mid h)=p_\star(\cdot\mid h)$.
\end{proposition}
\begin{proof}
The expected negative log-likelihood equals
$H(p_\star(\cdot\mid h))+D_{\mathrm{KL}}(p_\star(\cdot\mid h)\|q(\cdot\mid h))$.
The divergence is nonnegative and vanishes exactly when the 2 distributions coincide.
\end{proof}

\begin{proposition}[Effect of splitting a patch on its mean entropy]
\label{prop:segmentation_optimality}
For fixed per-position entropies $h_t$, let a patch of length $n$ have mean entropy $\mu$. Splitting it into two nonempty patches of lengths $n_1,n_2$ and means $\mu_1,\mu_2$ preserves total entropy. At least one child mean is at most $\mu$. The smaller child mean is strictly below $\mu$ if and only if the two child means differ.
\end{proposition}
\begin{proof}
The sum over positions gives $n\mu=n_1\mu_1+n_2\mu_2$, where $n=n_1+n_2$. Thus $\mu$ is a weighted average of the two child means and lies strictly between them when they differ.
\end{proof}

Proposition~\ref{prop:entropy_lower_bound} motivates predictive entropy as a score for next-token uncertainty. \texttt{EntroPE} estimates this score from a learned distribution. Proposition~\ref{prop:segmentation_optimality} describes what a split alone implies. The within-patch variance diagnostics and forecasting comparisons evaluate whether the selected boundaries are useful in practice.

\begin{remark}[Continuous forecasting outputs]
\label{rem:quantization}
The patch encoder embeds quantized input tokens. Its forecasting head projects the resulting hidden representations to real-valued predictions, so the output range is independent of the input bin centers.
\end{remark}

\FloatBarrier

\section{Forecasting Configurations and Results}
\label{app:extended_results}
\label{sec:rb:forecasting}

\subsection{Main Comparison and Training Protocol}
\label{app:main_forecast_protocol}

For the main comparison, we use $L=96$ and $T\in\{96,192,336,720\}$. Table~\ref{tab:fixed_hyperparams} gives the shared settings. Table~\ref{tab:exp_hyperparams} gives the configuration selected on the validation set for each dataset and horizon. For these configurations, we report the mean over three training seeds. Experiments used NVIDIA RTX 4090, RTX A5000, GeForce RTX 4090 D, and RTX 6000 Ada-16Q GPUs. We describe the separate seed experiments and historical results in Section~\ref{app:seed_analyses}.

For rerun baselines, we use the original authors' configurations, match the training budgets (epochs, early stopping, and timeout), and report the three-seed mean of the task metric. Forecasting uses MSE and MAE, classification uses accuracy, and anomaly detection uses F1.

\begin{table}[!htbp]
\centering
\caption{Shared settings for the main forecasting experiment at $L=96$.}
\label{tab:fixed_hyperparams}
\begin{tabular}{lc}
\toprule[1pt]
\textbf{Hyperparameter} & \textbf{Value} \\
\midrule
Input           & 96  \\
Global transformer layers       & 2 \\
Patch encoder / fusion decoder layers & 1 / 1 \\
Attention heads                 & 4 \\
FFN dimension ($d_{ff}$)        & 256 \\
Entropy vocabulary size         & 256 \\
Normalization                   & RevIN (affine) \\
LR schedule                     & One-cycle (pct\_start 0.3) \\
\bottomrule[1pt]
\end{tabular}
\end{table}

\begin{table*}[!htbp]
\centering
\scriptsize
\caption{Configurations selected for the main forecasting experiment. Small, medium, and large models use local/global dimensions of $(8,32)$, $(16,64)$, and $(32,128)$, respectively. $\alpha$ is the per-sample quantile level. Mono.\ selects entropy levels (0) or entropy increases (1).}
\label{tab:exp_hyperparams}
\begin{tabular}{llcccccc}
\toprule[1pt]
\textbf{Dataset} & $T$ & \textbf{Size} & \textbf{Batch} & \textbf{LR} & $\alpha$ & \textbf{Dropout} & \textbf{Mono.}  \\
\midrule
\multirow{4}{*}{ETTh1}   & 96  & M & 64  & $10^{-2}$ & 0.95 & 0.1 & 0 \\
                         & 192 & S & 128 & $10^{-2}$ & 0.75 & 0.1 & 1 \\
                         & 336 & S & 64  & $10^{-2}$ & 0.85 & 0.1 & 0 \\
                         & 720 & S & 32  & $10^{-2}$ & 0.85 & 0.2 & 0 \\
\midrule
\multirow{4}{*}{ETTh2}   & 96  & M & 32  & $10^{-2}$ & 0.85 & 0.3 & 1 \\
                         & 192 & L & 128 & $10^{-2}$ & 0.95 & 0.2 & 0 \\
                         & 336 & M & 128 & $10^{-2}$ & 0.95 & 0.3 & 0 \\
                         & 720 & M & 128 & $10^{-4}$ & 0.75 & 0.3 & 0 \\
\midrule
\multirow{4}{*}{ETTm1}   & 96  & S & 32  & $10^{-2}$ & 0.85 & 0.1 & 0 \\
                         & 192 & S & 128 & $10^{-2}$ & 0.95 & 0.1 & 1 \\
                         & 336 & S & 128 & $10^{-2}$ & 0.95 & 0.1 & 1 \\
                         & 720 & S & 32  & $10^{-2}$ & 0.85 & 0.1 & 1 \\
\midrule
\multirow{4}{*}{ETTm2}   & 96  & S & 64  & $10^{-2}$ & 0.75 & 0.2 & 1 \\
                         & 192 & M & 32  & $10^{-2}$ & 0.75 & 0.3 & 1 \\
                         & 336 & S & 32  & $10^{-2}$ & 0.75 & 0.2 & 0 \\
                         & 720 & L & 32  & $10^{-3}$ & 0.95 & 0.2 & 0 \\
\midrule
\multirow{4}{*}{Weather} & 96  & S & 32  & $10^{-2}$ & 0.95 & 0.2 & 1 \\
                         & 192 & S & 32  & $10^{-2}$ & 0.85 & 0.1 & 0 \\
                         & 336 & S & 32  & $10^{-2}$ & 0.95 & 0.2 & 1 \\
                         & 720 & S & 128 & $10^{-2}$ & 0.85 & 0.1 & 0 \\
\midrule
\multirow{4}{*}{ECL}     & 96  & M & 64  & $10^{-2}$ & 0.95 & 0.1 & 0 \\
                         & 192 & M & 32  & $10^{-2}$ & 0.85 & 0.1 & 0 \\
                         & 336 & M & 64  & $10^{-2}$ & 0.95 & 0.1 & 0 \\
                         & 720 & M & 64  & $10^{-2}$ & 0.95 & 0.1 & 0 \\
\midrule
\multirow{4}{*}{Traffic} & 96  & L & 8   & $10^{-2}$ & 0.95 & 0.1 & 1 \\
                         & 192 & L & 8   & $10^{-2}$ & 0.95 & 0.1 & 1 \\
                         & 336 & L & 8   & $10^{-2}$ & 0.95 & 0.1 & 1 \\
                         & 720 & L & 8   & $10^{-2}$ & 0.95 & 0.1 & 1 \\
\midrule
\multirow{4}{*}{Solar}   & 96  & M & 64  & $10^{-2}$ & 0.95 & 0.1 & 0 \\
                         & 192 & M & 32  & $10^{-2}$ & 0.85 & 0.1 & 0 \\
                         & 336 & M & 64  & $10^{-2}$ & 0.95 & 0.1 & 0 \\
                         & 720 & M & 32  & $10^{-2}$ & 0.85 & 0.1 & 1 \\
\bottomrule[1pt]
\end{tabular}
\end{table*}

\FloatBarrier

\subsection{Forecasting Errors by Dataset and Horizon}
\label{sec:full_forecast_table}

Table~\ref{tab:full_results} reports the individual horizon results underlying the main forecasting comparison. Within each dataset, the average row gives the mean over the four horizons. The final column averages across the eight datasets.

\begin{table*}[!htbp]
\centering
\caption{Full long-term forecasting results across all prediction horizons $T \in \{96, 192, 336, 720\}$ and their average (Avg.). Lower is better. \capbest{Best} and \capsecond{second-best} results are highlighted. ``All Dataset'' is the mean over the eight datasets. Ties are broken by the other metric of the same dataset.}
\label{tab:full_results}
\resizebox{\textwidth}{!}{%
\begin{tabular}{llcccccccccccccccccc}
\toprule
\multirow{2}{*}{Models} & \multirow{2}{*}{Horizon} & \multicolumn{2}{c}{ETTh1} & \multicolumn{2}{c}{ETTh2} & \multicolumn{2}{c}{ETTm1} & \multicolumn{2}{c}{ETTm2} & \multicolumn{2}{c}{Weather} & \multicolumn{2}{c}{Solar} & \multicolumn{2}{c}{ECL} & \multicolumn{2}{c}{Traffic} & \multicolumn{2}{c}{All Dataset} \\
\cmidrule(lr){3-4} \cmidrule(lr){5-6} \cmidrule(lr){7-8} \cmidrule(lr){9-10} \cmidrule(lr){11-12} \cmidrule(lr){13-14} \cmidrule(lr){15-16} \cmidrule(lr){17-18} \cmidrule(lr){19-20}
 & & MSE & MAE & MSE & MAE & MSE & MAE & MSE & MAE & MSE & MAE & MSE & MAE & MSE & MAE & MSE & MAE & MSE & MAE \\
\midrule
\multirow{5}{*}{FilterTS} & 96 & 0.388 & 0.396 & 0.292 & 0.341 & 0.325 & 0.362 & \best{0.174} & \second{0.256} & 0.173 & 0.217 & 0.230 & 0.272 & 0.174 & 0.261 & 0.466 & 0.319 & 0.278 & 0.303 \\
 & 192 & 0.436 & \second{0.423} & 0.378 & 0.392 & 0.367 & 0.384 & \best{0.240} & \second{0.299} & 0.223 & 0.260 & 0.268 & 0.296 & 0.184 & 0.270 & 0.472 & 0.318 & 0.321 & 0.330 \\
 & 336 & 0.479 & 0.446 & 0.423 & 0.429 & 0.400 & 0.405 & \best{0.303} & \second{0.340} & 0.279 & 0.300 & 0.298 & 0.314 & 0.199 & 0.285 & 0.488 & 0.325 & 0.359 & 0.356 \\
 & 720 & 0.482 & 0.472 & 0.429 & 0.444 & 0.464 & 0.441 & 0.409 & 0.400 & 0.356 & 0.350 & 0.299 & 0.314 & 0.241 & 0.319 & 0.520 & 0.343 & 0.400 & 0.385 \\
\cmidrule(lr){2-20}
 & Avg. & 0.446 & 0.434 & 0.381 & 0.402 & 0.389 & 0.398 & \best{0.282} & \second{0.324} & 0.258 & 0.282 & 0.274 & 0.299 & 0.200 & 0.284 & 0.487 & 0.326 & 0.339 & 0.344 \\
\midrule
\multirow{5}{*}{HDMixer} & 96 & 0.388 & 0.398 & 0.293 & 0.343 & 0.336 & 0.368 & 0.182 & 0.265 & 0.174 & 0.224 & 0.256 & 0.319 & 0.175 & 0.279 & 0.525 & 0.373 & 0.291 & 0.321 \\
 & 192 & 0.443 & 0.429 & 0.379 & 0.395 & 0.375 & 0.386 & 0.246 & 0.306 & 0.225 & 0.264 & 0.322 & 0.328 & 0.187 & 0.291 & 0.519 & 0.375 & 0.337 & 0.347 \\
 & 336 & 0.483 & 0.449 & 0.420 & 0.431 & 0.406 & 0.407 & 0.305 & 0.343 & 0.274 & 0.302 & 0.336 & 0.350 & 0.198 & 0.303 & 0.549 & 0.381 & 0.371 & 0.371 \\
 & 720 & 0.495 & 0.477 & 0.438 & 0.450 & 0.474 & 0.445 & 0.402 & 0.400 & 0.351 & 0.348 & 0.336 & 0.346 & 0.236 & 0.330 & 0.579 & 0.387 & 0.414 & 0.398 \\
\cmidrule(lr){2-20}
 & Avg. & 0.452 & 0.438 & 0.383 & 0.405 & 0.398 & 0.402 & 0.284 & 0.329 & 0.256 & 0.285 & 0.313 & 0.336 & 0.199 & 0.301 & 0.543 & 0.379 & 0.353 & 0.359 \\
\midrule
\multirow{5}{*}{TimeBase} & 96 & 0.394 & 0.402 & 0.301 & 0.348 & 0.386 & 0.408 & 0.195 & 0.278 & 0.173 & 0.220 & 0.355 & 0.389 & 0.222 & 0.315 & 0.767 & 0.404 & 0.349 & 0.346 \\
 & 192 & 0.442 & 0.429 & 0.396 & 0.409 & 0.410 & 0.413 & 0.253 & 0.313 & 0.220 & 0.262 & 0.395 & 0.413 & 0.229 & 0.323 & 0.696 & 0.381 & 0.380 & 0.368 \\
 & 336 & 0.489 & 0.457 & 0.433 & 0.441 & 0.452 & 0.444 & 0.315 & 0.351 & 0.274 & 0.302 & 0.431 & 0.427 & 0.247 & 0.339 & 0.705 & 0.383 & 0.418 & 0.393 \\
 & 720 & 0.475 & 0.465 & 0.444 & 0.460 & 0.503 & 0.464 & 0.418 & 0.409 & 0.356 & 0.356 & 0.428 & 0.416 & 0.290 & 0.371 & 0.746 & 0.401 & 0.458 & 0.418 \\
\cmidrule(lr){2-20}
 & Avg. & 0.450 & 0.438 & 0.394 & 0.415 & 0.438 & 0.432 & 0.295 & 0.338 & 0.256 & 0.285 & 0.402 & 0.411 & 0.247 & 0.337 & 0.729 & 0.392 & 0.401 & 0.381 \\
\midrule
\multirow{5}{*}{TimeMosaic} & 96 & \best{0.375} & \best{0.387} & \second{0.291} & \best{0.332} & \best{0.311} & \best{0.339} & \second{0.177} & \best{0.252} & 0.175 & \best{0.208} & 0.231 & \second{0.246} & 0.176 & 0.253 & 0.456 & 0.279 & 0.274 & \second{0.287} \\
 & 192 & 0.424 & \best{0.416} & \best{0.368} & \best{0.383} & 0.368 & \best{0.369} & 0.245 & \best{0.296} & 0.222 & \best{0.250} & 0.268 & 0.272 & 0.182 & 0.261 & 0.469 & 0.282 & 0.318 & \second{0.316} \\
 & 336 & 0.473 & \best{0.440} & 0.420 & \second{0.420} & 0.413 & \best{0.397} & 0.306 & \best{0.336} & 0.277 & \second{0.291} & 0.293 & 0.281 & 0.199 & 0.278 & 0.488 & 0.291 & 0.359 & \best{0.342} \\
 & 720 & \second{0.474} & \second{0.461} & \second{0.419} & \best{0.436} & 0.465 & \best{0.430} & \best{0.402} & \best{0.393} & 0.354 & \best{0.341} & 0.281 & \second{0.275} & 0.239 & 0.310 & 0.525 & \second{0.307} & \second{0.395} & \best{0.369} \\
\cmidrule(lr){2-20}
 & Avg. & \second{0.437} & \best{0.426} & \second{0.375} & \best{0.393} & 0.389 & \best{0.384} & 0.283 & \best{0.319} & 0.257 & \best{0.273} & 0.268 & 0.269 & 0.199 & 0.276 & 0.485 & 0.290 & 0.336 & \second{0.329} \\
\midrule
\multirow{5}{*}{TSLANet} & 96 & 0.380 & 0.402 & 0.299 & 0.350 & 0.322 & 0.360 & 0.178 & 0.262 & 0.174 & 0.215 & 0.232 & 0.267 & 0.177 & 0.262 & 0.470 & 0.300 & 0.279 & 0.302 \\
 & 192 & 0.437 & 0.434 & 0.380 & 0.401 & \best{0.361} & 0.382 & 0.249 & 0.308 & 0.221 & 0.256 & 0.255 & 0.280 & 0.183 & 0.268 & 0.477 & 0.300 & 0.320 & 0.329 \\
 & 336 & 0.488 & 0.456 & 0.418 & 0.426 & 0.394 & 0.406 & 0.310 & 0.347 & 0.276 & 0.297 & 0.281 & 0.294 & 0.200 & 0.286 & 0.488 & 0.305 & 0.357 & 0.352 \\
 & 720 & 0.550 & 0.504 & 0.427 & 0.445 & \second{0.456} & 0.442 & 0.408 & 0.404 & 0.352 & 0.346 & 0.284 & 0.293 & 0.242 & 0.320 & 0.519 & 0.323 & 0.405 & 0.385 \\
\cmidrule(lr){2-20}
 & Avg. & 0.464 & 0.449 & 0.381 & 0.406 & \second{0.383} & 0.398 & 0.286 & 0.330 & 0.256 & 0.279 & 0.263 & 0.284 & 0.201 & 0.284 & 0.489 & 0.307 & 0.340 & 0.342 \\
\midrule
\multirow{5}{*}{CALF} & 96 & 0.378 & \second{0.394} & 0.297 & 0.342 & 0.327 & 0.355 & 0.180 & 0.256 & 0.175 & 0.211 & \best{0.202} & \best{0.220} & 0.161 & \best{0.243} & \second{0.431} & \best{0.268} & \best{0.269} & \best{0.286} \\
 & 192 & 0.430 & 0.424 & 0.376 & \second{0.391} & 0.376 & 0.381 & 0.246 & 0.300 & \second{0.220} & 0.254 & \best{0.232} & \best{0.238} & 0.176 & \second{0.257} & \second{0.453} & \best{0.277} & \best{0.314} & \best{0.315} \\
 & 336 & 0.480 & 0.450 & 0.418 & 0.431 & 0.425 & 0.407 & 0.307 & 0.341 & 0.277 & 0.294 & 0.262 & \best{0.257} & 0.194 & \second{0.275} & \second{0.476} & \best{0.287} & \second{0.355} & \second{0.343} \\
 & 720 & 0.484 & 0.472 & 0.424 & 0.442 & 0.483 & 0.445 & 0.406 & 0.398 & 0.355 & 0.347 & 0.271 & \best{0.257} & 0.235 & \second{0.309} & \second{0.507} & \best{0.306} & 0.396 & \second{0.372} \\
\cmidrule(lr){2-20}
 & Avg. & 0.443 & 0.435 & 0.379 & 0.402 & 0.403 & 0.397 & 0.285 & 0.324 & 0.257 & 0.277 & \second{0.242} & \best{0.243} & 0.192 & \second{0.271} & \second{0.467} & \best{0.285} & \second{0.333} & \best{0.329} \\
\midrule
\multirow{5}{*}{DLinear} & 96 & 0.398 & 0.412 & 0.351 & 0.403 & 0.346 & 0.374 & 0.195 & 0.295 & 0.197 & 0.259 & 0.287 & 0.374 & 0.196 & 0.281 & 0.652 & 0.400 & 0.328 & 0.350 \\
 & 192 & 0.446 & 0.441 & 0.482 & 0.479 & 0.382 & 0.391 & 0.285 & 0.361 & 0.239 & 0.299 & 0.317 & 0.395 & 0.196 & 0.284 & 0.601 & 0.375 & 0.369 & 0.378 \\
 & 336 & 0.487 & 0.465 & 0.602 & 0.546 & 0.415 & 0.415 & 0.385 & 0.429 & 0.282 & 0.332 & 0.351 & 0.413 & 0.209 & 0.300 & 0.608 & 0.377 & 0.417 & 0.410 \\
 & 720 & 0.516 & 0.513 & 0.840 & 0.661 & 0.474 & 0.451 & 0.556 & 0.523 & 0.347 & 0.384 & 0.356 & 0.413 & 0.244 & 0.332 & 0.648 & 0.398 & 0.498 & 0.459 \\
\cmidrule(lr){2-20}
 & Avg. & 0.462 & 0.458 & 0.569 & 0.522 & 0.404 & 0.408 & 0.355 & 0.402 & 0.266 & 0.319 & 0.328 & 0.399 & 0.211 & 0.299 & 0.627 & 0.388 & 0.403 & 0.399 \\
\midrule
\multirow{5}{*}{FEDformer} & 96 & \second{0.377} & 0.419 & 0.351 & 0.392 & 0.385 & 0.422 & 0.195 & 0.284 & 0.222 & 0.304 & 0.236 & 0.355 & 0.194 & 0.311 & 0.566 & 0.348 & 0.316 & 0.354 \\
 & 192 & \best{0.421} & 0.445 & 0.442 & 0.450 & 0.441 & 0.448 & 0.266 & 0.326 & 0.299 & 0.360 & 0.255 & 0.359 & 0.212 & 0.328 & 0.607 & 0.381 & 0.368 & 0.387 \\
 & 336 & \second{0.469} & 0.473 & 0.499 & 0.490 & 0.466 & 0.470 & 0.331 & 0.368 & 0.339 & 0.386 & \best{0.247} & 0.355 & 0.219 & 0.335 & 0.609 & 0.370 & 0.397 & 0.406 \\
 & 720 & 0.504 & 0.503 & 0.493 & 0.494 & 0.529 & 0.498 & 0.438 & 0.433 & 0.421 & 0.435 & 0.278 & 0.383 & 0.247 & 0.356 & 0.643 & 0.397 & 0.444 & 0.437 \\
\cmidrule(lr){2-20}
 & Avg. & 0.443 & 0.460 & 0.446 & 0.457 & 0.455 & 0.460 & 0.308 & 0.353 & 0.320 & 0.371 & 0.254 & 0.363 & 0.218 & 0.333 & 0.606 & 0.374 & 0.381 & 0.396 \\
\midrule
\multirow{5}{*}{PatchTST} & 96 & 0.383 & 0.405 & 0.314 & 0.358 & 0.336 & 0.374 & 0.183 & 0.266 & 0.174 & 0.216 & 0.218 & 0.275 & 0.176 & 0.270 & 0.454 & 0.293 & 0.280 & 0.307 \\
 & 192 & 0.447 & 0.440 & 0.391 & 0.407 & 0.372 & 0.394 & 0.255 & 0.315 & 0.220 & 0.256 & 0.250 & 0.295 & 0.184 & 0.277 & 0.461 & 0.296 & 0.323 & 0.335 \\
 & 336 & 0.490 & 0.472 & 0.431 & 0.441 & 0.410 & 0.417 & 0.321 & 0.356 & 0.276 & 0.296 & 0.274 & 0.308 & 0.201 & 0.294 & 0.477 & 0.303 & 0.360 & 0.361 \\
 & 720 & 0.525 & 0.503 & 0.442 & 0.457 & 0.465 & 0.452 & 0.416 & 0.414 & 0.353 & 0.346 & 0.267 & 0.299 & 0.241 & 0.326 & 0.510 & 0.322 & 0.402 & 0.390 \\
\cmidrule(lr){2-20}
 & Avg. & 0.461 & 0.455 & 0.395 & 0.416 & 0.396 & 0.409 & 0.294 & 0.338 & 0.256 & 0.279 & 0.252 & 0.294 & 0.201 & 0.292 & 0.476 & 0.304 & 0.341 & 0.348 \\
\midrule
\multirow{5}{*}{TimeMixer} & 96 & 0.394 & 0.402 & 0.295 & 0.344 & 0.327 & 0.364 & 0.179 & 0.261 & \best{0.163} & \second{0.210} & 0.206 & 0.257 & \second{0.157} & 0.248 & 0.472 & 0.301 & 0.274 & 0.298 \\
 & 192 & 0.453 & 0.439 & 0.390 & 0.403 & 0.374 & 0.386 & \second{0.242} & 0.302 & 0.225 & 0.266 & \second{0.233} & 0.275 & \second{0.171} & 0.261 & 0.481 & 0.304 & 0.321 & 0.330 \\
 & 336 & 0.498 & 0.455 & 0.437 & 0.443 & \second{0.392} & 0.406 & \second{0.303} & 0.342 & \best{0.264} & \best{0.291} & 0.258 & 0.297 & \second{0.190} & 0.282 & 0.504 & 0.309 & 0.356 & 0.353 \\
 & 720 & 0.585 & 0.521 & 0.466 & 0.460 & 0.464 & 0.447 & \second{0.402} & \second{0.397} & \second{0.346} & 0.346 & 0.301 & 0.317 & \second{0.227} & 0.312 & 0.521 & 0.327 & 0.414 & 0.391 \\
\cmidrule(lr){2-20}
 & Avg. & 0.483 & 0.454 & 0.397 & 0.413 & 0.389 & 0.401 & \second{0.282} & 0.326 & \second{0.250} & 0.278 & 0.250 & 0.287 & \second{0.186} & 0.276 & 0.495 & 0.310 & 0.341 & 0.343 \\
\midrule
\multirow{5}{*}{iTransformer} & 96 & 0.391 & 0.406 & 0.305 & 0.355 & 0.346 & 0.377 & 0.184 & 0.266 & 0.177 & 0.216 & 0.218 & 0.247 & \best{0.154} & \second{0.244} & \best{0.407} & \second{0.273} & \second{0.273} & 0.298 \\
 & 192 & 0.448 & 0.441 & 0.381 & 0.404 & 0.398 & 0.405 & 0.252 & 0.310 & 0.224 & 0.257 & 0.239 & \second{0.266} & \best{0.167} & \best{0.257} & \best{0.431} & \second{0.282} & 0.318 & 0.328 \\
 & 336 & 0.485 & 0.456 & 0.436 & 0.436 & 0.439 & 0.429 & 0.316 & 0.353 & 0.282 & 0.299 & 0.256 & \second{0.279} & \best{0.184} & \best{0.275} & \best{0.449} & \second{0.291} & 0.356 & 0.352 \\
 & 720 & 0.504 & 0.491 & 0.462 & 0.461 & 0.501 & 0.466 & 0.419 & 0.411 & 0.359 & 0.351 & \second{0.261} & 0.281 & \best{0.221} & \best{0.307} & \best{0.481} & 0.311 & 0.401 & 0.385 \\
\cmidrule(lr){2-20}
 & Avg. & 0.457 & 0.449 & 0.396 & 0.414 & 0.421 & 0.419 & 0.293 & 0.335 & 0.261 & 0.281 & 0.244 & \second{0.268} & \best{0.182} & \best{0.271} & \best{0.442} & \second{0.289} & 0.337 & 0.341 \\
\midrule
\multirow{5}{*}{TimesNet} & 96 & 0.415 & 0.431 & 0.338 & 0.379 & 0.336 & 0.377 & 0.188 & 0.268 & 0.178 & 0.226 & 0.297 & 0.328 & 0.208 & 0.310 & 0.628 & 0.350 & 0.324 & 0.334 \\
 & 192 & 0.464 & 0.456 & 0.406 & 0.414 & 0.388 & 0.402 & 0.256 & 0.310 & 0.225 & 0.265 & 0.381 & 0.377 & 0.226 & 0.324 & 0.639 & 0.350 & 0.373 & 0.362 \\
 & 336 & 0.498 & 0.472 & 0.471 & 0.468 & 0.417 & 0.425 & 0.320 & 0.349 & 0.287 & 0.308 & 0.341 & 0.373 & 0.268 & 0.350 & 0.655 & 0.359 & 0.407 & 0.388 \\
 & 720 & 0.529 & 0.502 & 0.441 & 0.452 & 0.473 & 0.452 & 0.429 & 0.411 & 0.357 & 0.353 & 0.360 & 0.381 & 0.302 & 0.373 & 0.687 & 0.372 & 0.447 & 0.412 \\
\cmidrule(lr){2-20}
 & Avg. & 0.477 & 0.465 & 0.414 & 0.428 & 0.404 & 0.414 & 0.298 & 0.335 & 0.262 & 0.288 & 0.345 & 0.365 & 0.251 & 0.339 & 0.652 & 0.358 & 0.388 & 0.374 \\
\midrule
\multirow{5}{*}{TimeFFM} & 96 & 0.400 & 0.403 & \best{0.290} & \second{0.341} & 0.338 & 0.362 & 0.191 & 0.276 & 0.195 & 0.237 & 0.328 & 0.370 & 0.200 & 0.283 & 0.691 & 0.440 & 0.329 & 0.339 \\
 & 192 & 0.457 & 0.436 & \second{0.371} & 0.393 & 0.380 & 0.383 & 0.247 & 0.307 & 0.240 & 0.272 & 0.368 & 0.381 & 0.212 & 0.304 & 0.647 & 0.424 & 0.365 & 0.363 \\
 & 336 & 0.497 & 0.453 & \second{0.414} & 0.428 & 0.413 & 0.404 & 0.313 & 0.351 & 0.292 & 0.307 & 0.401 & 0.396 & 0.225 & 0.314 & 0.651 & 0.425 & 0.401 & 0.385 \\
 & 720 & 0.483 & 0.463 & 0.425 & 0.444 & 0.477 & 0.438 & 0.409 & 0.405 & 0.364 & 0.352 & 0.401 & 0.397 & 0.258 & 0.337 & 0.687 & 0.440 & 0.438 & 0.410 \\
\cmidrule(lr){2-20}
 & Avg. & 0.459 & 0.439 & 0.375 & 0.402 & 0.402 & 0.397 & 0.290 & 0.335 & 0.273 & 0.292 & 0.375 & 0.386 & 0.224 & 0.310 & 0.669 & 0.432 & 0.383 & 0.374 \\
\midrule
\multirow{5}{*}{\textbf{\oursbold{}}} & 96 & 0.379 & 0.396 & 0.295 & 0.348 & \second{0.315} & \second{0.355} & 0.180 & 0.267 & \second{0.168} & 0.216 & \second{0.206} & 0.254 & 0.169 & 0.262 & 0.485 & 0.288 & 0.275 & 0.298 \\
 & 192 & \second{0.422} & 0.425 & 0.378 & 0.399 & \second{0.362} & \second{0.381} & 0.246 & 0.306 & \best{0.212} & \second{0.254} & 0.240 & 0.276 & 0.178 & 0.269 & 0.488 & 0.295 & \second{0.316} & 0.326 \\
 & 336 & \best{0.459} & \second{0.446} & \best{0.387} & \best{0.420} & \best{0.388} & \second{0.402} & 0.309 & 0.348 & \second{0.267} & 0.295 & \second{0.255} & 0.286 & 0.194 & 0.284 & 0.494 & 0.297 & \best{0.344} & 0.347 \\
 & 720 & \best{0.446} & \best{0.458} & \best{0.414} & \second{0.439} & \best{0.449} & \second{0.436} & 0.406 & 0.406 & \best{0.341} & \second{0.342} & \best{0.255} & 0.283 & 0.232 & 0.320 & 0.525 & 0.311 & \best{0.384} & 0.374 \\
\cmidrule(lr){2-20}
 & Avg. & \best{0.427} & \second{0.431} & \best{0.369} & \second{0.402} & \best{0.379} & \second{0.394} & 0.285 & 0.332 & \best{0.247} & \second{0.277} & \best{0.239} & 0.275 & 0.193 & 0.284 & 0.498 & 0.298 & \best{0.330} & 0.336 \\
\bottomrule
\end{tabular}}
\end{table*}

\FloatBarrier

\subsection{Longer Contexts and Forecast Horizons}
\label{sec:rb:long_context}

We compare dynamic and fixed patching at small, medium, and large model sizes. Within each comparison, the training budget is held fixed and the static patch length matches the average dynamic patch length. The static comparator follows the PatchTST architecture. TimeMosaic provides an additional adaptive-patching baseline. The ETT, Weather, and Solar comparisons use $L=512$. M4-Monthly uses $L=36$ and $T=18$.

Table~\ref{tab:rb:long_context} gives absolute errors, and Table~\ref{tab:rb:relative_changes} expresses each paired difference relative to the static model. At $512\!\to\!96$, dynamic patching has lower MSE for all five datasets and all three sizes. Gains are mixed for M4-Monthly and at longer horizons. On Solar, static patching has lower MSE for the small model at $T=480$ and $T=720$ and for the large model at $T=720$.

\begin{table*}[!htbp]
\centering
\caption{Forecasting results for short-, medium-, and long-horizon settings.
In each S/M/L column, entries are \textbf{Dynamic/Static} for the indicated metric. The final column instead reports \textbf{MSE/MAE} for TimeMosaic. Lower error is better. S/M/L denote model size, while $L$ in $L\!\to\!T$ denotes input length.}
\label{tab:rb:long_context}
\begingroup
\scriptsize
\setlength{\tabcolsep}{3pt}
\renewcommand{\arraystretch}{1.12}
\resizebox{\textwidth}{!}{%
\begin{tabular}{lcccccccc}
\toprule
Dataset & L $\!\to\!$ T & S-MSE & S-MAE & M-MSE & M-MAE & L-MSE & L-MAE & TimeMosaic \\
 & & & & & & & & (MSE/MAE) \\
\midrule
M4-Monthly & 36 $\!\to\!$ 18 & .200/.199 & .195/.197 & .196/.199 & .193/.196 & .196/.195 & .192/.192 & .198/.194 \\
ETTh1   & 512 $\!\to\!$ 96 & .392/.420 & .417/.437 & .408/.422 & .427/.438 & .387/.443 & .410/.452 & .389/.418 \\
ETTh2   & 512 $\!\to\!$ 96 & .312/.318 & .371/.380 & .294/.324 & .357/.384 & .322/.334 & .375/.380 & .314/.318 \\
ETTm1   & 512 $\!\to\!$ 96 & .310/.326 & .357/.370 & .324/.337 & .364/.381 & .305/.313 & .353/.361 & .320/.362 \\
ETTm2   & 512 $\!\to\!$ 96 & .182/.188 & .270/.276 & .178/.194 & .271/.277 & .179/.194 & .267/.279 & .179/.279 \\
Weather & 512 $\!\to\!$ 96 & .159/.172 & .215/.230 & .158/.166 & .214/.221 & .159/.163 & .212/.217 & .156/.220 \\
\midrule
ETTh1   & 512 $\!\to\!$ 480 & .451/.485 & .463/.485 & .434/.474 & .457/.484 & .432/.463 & .452/.478 & .435/.455 \\
ETTh2   & 512 $\!\to\!$ 480 & .380/.380 & .431/.432 & .367/.386 & .416/.432 & .399/.393 & .442/.437 & .361/.415 \\
ETTm1   & 512 $\!\to\!$ 480 & .394/.412 & .407/.417 & .430/.424 & .430/.423 & .397/.423 & .406/.425 & .391/.419 \\
ETTm2   & 512 $\!\to\!$ 480 & .349/.349 & .380/.383 & .343/.360 & .381/.387 & .331/.346 & .367/.384 & .348/.380 \\
Solar   & 512 $\!\to\!$ 480 & .260/.245 & .331/.315 & .222/.250 & .300/.321 & .233/.233 & .315/.318 & .252/.300 \\
Weather & 512 $\!\to\!$ 480 & .286/.290 & .312/.313 & .291/.290 & .316/.316 & .286/.288 & .309/.311 & .288/.316 \\
\midrule
Solar   & 512 $\!\to\!$ 720 & .255/.250 & .320/.319 & .229/.249 & .301/.323 & .243/.237 & .318/.318 & .256/.319 \\
\bottomrule
\end{tabular}}
\endgroup
\end{table*}

\begin{table*}[!htbp]
\centering
\caption{Relative error reduction (\%) for dynamic versus static patching in Table~\ref{tab:rb:long_context}, defined as $100(e_{\mathrm{static}}-e_{\mathrm{dynamic}})/e_{\mathrm{static}}$. Positive values favor dynamic patching; negative values favor static patching. 
}
\label{tab:rb:relative_changes}
\begingroup
\scriptsize
\setlength{\tabcolsep}{6pt}
\renewcommand{\arraystretch}{1.12}
\resizebox{\textwidth}{!}{%
\begin{tabular}{lccccccc}
\toprule
Dataset     & L $\!\to\!$ T & S-MSE & S-MAE & M-MSE & M-MAE & L-MSE & L-MAE \\
\midrule
M4-Monthly  & 36 $\!\to\!$ 18 & -0.313 & +0.770 & +1.336 & +1.317 & -0.465 & -0.346 \\
ETTh1       & 512 $\!\to\!$ 96 & +6.845 & +4.467 & +3.201 & +2.490 & +12.654 & +9.340 \\
ETTh2       & 512 $\!\to\!$ 96 & +1.945 & +2.530 & +9.462 & +6.998 & +3.591 & +1.264 \\
ETTm1       & 512 $\!\to\!$ 96 & +5.074 & +3.594 & +3.783 & +4.561 & +2.432 & +2.164 \\
ETTm2       & 512 $\!\to\!$ 96 & +3.276 & +2.272 & +8.108 & +1.976 & +7.700 & +4.310 \\
Weather     & 512 $\!\to\!$ 96 & +7.360 & +6.868 & +4.876 & +2.968 & +2.086 & +2.229 \\
\midrule
ETTh1       & 512 $\!\to\!$ 480 & +6.958 & +4.416 & +8.394 & +5.682 & +6.789 & +5.440 \\
ETTh2       & 512 $\!\to\!$ 480 & -0.027 & +0.407 & +4.889 & +3.836 & -1.514 & -1.240 \\
ETTm1       & 512 $\!\to\!$ 480 & +4.418 & +2.576 & -1.303 & -1.489 & +6.254 & +4.508 \\
ETTm2       & 512 $\!\to\!$ 480 & -0.233 & +0.909 & +4.892 & +1.439 & +4.353 & +4.494 \\
Solar       & 512 $\!\to\!$ 480 & -6.053 & -5.012 & +11.155 & +6.473 & +0.105 & +0.743 \\
Weather     & 512 $\!\to\!$ 480 & +1.423 & +0.258 & -0.183 & +0.046 & +0.733 & +0.523 \\
\midrule
Solar       & 512 $\!\to\!$ 720 & -2.038 & -0.206 & +8.130 & +6.732 & -2.336 & +0.048 \\
\bottomrule
\end{tabular}}
\endgroup
\end{table*}

\FloatBarrier

\subsection{PatchTST Integration Experiment}
\label{app:integration}

We replace PatchTST's static patching with EntroPE's dynamic patcher and fusion decoder while keeping the remaining components fixed. Table~\ref{tab:patchtst_integration} reports this six-dataset integration experiment separately from the eight-dataset main comparison. Both changed components contribute to the comparison. The combined variant has lower MSE on all six datasets. On ETTm2, MAE changes from 0.334 to 0.335.

\begin{table}[!htbp]
\centering
\caption{PatchTST integration experiment on six datasets. The second row uses EntroPE's dynamic patcher and fusion decoder. Relative reductions use PatchTST as the reference and are rounded to whole percentages.}
\label{tab:patchtst_integration}
\resizebox{\textwidth}{!}{
\begin{tabular}{lcccccccccccc}
\toprule
\textbf{Model} & \multicolumn{2}{c}{\textbf{ETTh1}} & \multicolumn{2}{c}{\textbf{ETTh2}} & \multicolumn{2}{c}{\textbf{ETTm1}} & \multicolumn{2}{c}{\textbf{ETTm2}} & \multicolumn{2}{c}{\textbf{Weather}} & \multicolumn{2}{c}{\textbf{ECL}} \\
& MSE & MAE & MSE & MAE & MSE & MAE & MSE & MAE & MSE & MAE & MSE & MAE \\
\midrule
PatchTST & 0.516 & 0.484 & 0.391 & 0.411 & 0.406 & 0.407 & 0.290 & 0.334 & 0.265 & 0.285 & 0.216 & 0.318 \\
\texttt{EntroPE} & 0.416 & 0.425 & 0.366 & 0.387 & 0.378 & 0.391 & 0.286 & 0.335 & 0.242 & 0.273 & 0.182 & 0.271 \\
\midrule
$\Delta$\% Improvement & 19\% & 12\% & 6\% & 6\% & 7\% & 4\% & 1\% & 0\% & 9\% & 4\% & 16\% & 15\% \\
\bottomrule
\end{tabular}
}
\end{table}

\FloatBarrier

\section{Forecasting Variability Across Seeds}
\label{app:seed_analyses}

Table~\ref{tab:forecast_seed_std} compares \texttt{EntroPE} with four baselines in the main forecasting evaluation. It reports horizon-averaged errors across three seeds on seven datasets and single-seed results on Traffic. We then examine \texttt{EntroPE} across five seeds in two settings. Table~\ref{tab:rb:shared_seeds} reports per-horizon variability under a fixed configuration. Finally, Table~\ref{tab:best_seed_full} reproduces the earlier six-dataset comparison, in which we evaluated \texttt{EntroPE} using best-seed reporting.

\subsection{Seed Variability in the Main Forecasting Comparison}
\label{app:main_forecast_seed_variability}

Table~\ref{tab:forecast_seed_std} compares \texttt{EntroPE} with four baselines from the main forecasting benchmark. Across the seven datasets evaluated with three seeds, \texttt{EntroPE}'s horizon-averaged MSE standard deviation ranges from 0.001 on ETTm2, Weather, and Electricity to 0.005 on ETTh2. Among the five methods shown, \texttt{EntroPE} has the lowest mean MSE on ETTh1, ETTh2, ETTm1, Weather, and Solar. TimeMosaic has the lowest mean MSE on ETTm2, and iTransformer on Electricity.

\begin{table}[!htbp]
\centering
\caption{Horizon-averaged test MSE and MAE at input length $L=96$. For each seed, we average each metric over $T\in\{96,192,336,720\}$. Entries give the mean $\pm$ sample standard deviation across seeds 2021, 2022, and 2023. Traffic entries give the single-seed score. A displayed standard deviation of $0.000$ is below $0.0005$.}
\label{tab:forecast_seed_std}
\label{tab:forecast_seed_std}
\scriptsize
\setlength{\tabcolsep}{3pt}
\renewcommand{\arraystretch}{1.08}
\begin{tabular}{@{}llccccc@{}}
\toprule
\textbf{Dataset} & \textbf{Metric} & \textbf{TimeMosaic} & \textbf{CALF}
& \textbf{iTransformer} & \textbf{HDMixer} & \textbf{EntroPE} \\
\midrule
ETTh1 & MSE & 0.437 $\pm$ 0.004 & 0.443 $\pm$ 0.002 & 0.457 $\pm$ 0.003 & 0.452 $\pm$ 0.003 & 0.427 $\pm$ 0.004 \\
 & MAE & 0.426 $\pm$ 0.003 & 0.435 $\pm$ 0.001 & 0.449 $\pm$ 0.002 & 0.438 $\pm$ 0.003 & 0.431 $\pm$ 0.002 \\
\addlinespace[2pt]
ETTh2 & MSE & 0.375 $\pm$ 0.002 & 0.379 $\pm$ 0.003 & 0.396 $\pm$ 0.005 & 0.383 $\pm$ 0.005 & 0.369 $\pm$ 0.005 \\
 & MAE & 0.393 $\pm$ 0.001 & 0.402 $\pm$ 0.001 & 0.414 $\pm$ 0.003 & 0.405 $\pm$ 0.001 & 0.402 $\pm$ 0.003 \\
\addlinespace[2pt]
ETTm1 & MSE & 0.389 $\pm$ 0.003 & 0.403 $\pm$ 0.001 & 0.421 $\pm$ 0.004 & 0.398 $\pm$ 0.002 & 0.379 $\pm$ 0.003 \\
 & MAE & 0.384 $\pm$ 0.001 & 0.397 $\pm$ 0.000 & 0.419 $\pm$ 0.003 & 0.402 $\pm$ 0.001 & 0.394 $\pm$ 0.001 \\
\addlinespace[2pt]
ETTm2 & MSE & 0.283 $\pm$ 0.001 & 0.285 $\pm$ 0.001 & 0.293 $\pm$ 0.002 & 0.284 $\pm$ 0.002 & 0.285 $\pm$ 0.001 \\
 & MAE & 0.319 $\pm$ 0.000 & 0.324 $\pm$ 0.000 & 0.335 $\pm$ 0.002 & 0.329 $\pm$ 0.002 & 0.332 $\pm$ 0.000 \\
\addlinespace[2pt]
Weather & MSE & 0.257 $\pm$ 0.000 & 0.257 $\pm$ 0.000 & 0.261 $\pm$ 0.001 & 0.256 $\pm$ 0.000 & 0.247 $\pm$ 0.001 \\
 & MAE & 0.273 $\pm$ 0.001 & 0.277 $\pm$ 0.000 & 0.281 $\pm$ 0.001 & 0.285 $\pm$ 0.000 & 0.277 $\pm$ 0.000 \\
\addlinespace[2pt]
Solar & MSE & 0.268 $\pm$ 0.001 & 0.242 $\pm$ 0.001 & 0.244 $\pm$ 0.002 & 0.313 $\pm$ 0.011 & 0.239 $\pm$ 0.003 \\
 & MAE & 0.269 $\pm$ 0.001 & 0.243 $\pm$ 0.001 & 0.268 $\pm$ 0.000 & 0.336 $\pm$ 0.001 & 0.275 $\pm$ 0.001 \\
\addlinespace[2pt]
ECL & MSE & 0.199 $\pm$ 0.000 & 0.192 $\pm$ 0.001 & 0.182 $\pm$ 0.000 & 0.199 $\pm$ 0.001 & 0.193 $\pm$ 0.001 \\
 & MAE & 0.276 $\pm$ 0.001 & 0.271 $\pm$ 0.001 & 0.271 $\pm$ 0.000 & 0.301 $\pm$ 0.001 & 0.284 $\pm$ 0.001 \\
\addlinespace[2pt]
Traffic & MSE & 0.485 & 0.467 & 0.442 & 0.543 & 0.498 \\
 & MAE & 0.290 & 0.285 & 0.289 & 0.379 & 0.298 \\
\bottomrule
\end{tabular}
\end{table}

\FloatBarrier

\subsection{Per-Horizon Seed Variability Under a Fixed Configuration}
\label{sec:rb:seed_robustness}

To measure how forecasting error varies with the training seed, we evaluate \texttt{EntroPE} on five datasets using seeds 2021--2025. Across datasets and horizons, we fix the model width ($d_{\mathrm{model}}=16$), depth (two layers), number of attention heads (four), entropy quantile ($\alpha=0.75$), and learning rate ($0.001$). Table~\ref{tab:rb:shared_seeds} reports the mean and standard deviation separately for each forecast horizon. The largest MSE standard deviations occur at $T=720$ on ETTh1 ($0.013$), ETTh2 ($0.012$), and ETTm2 ($0.010$). On Weather, the MSE standard deviation is at most $0.001$ at every horizon.

\begin{table}[!htbp]
\centering
\caption{Five-seed robustness under a shared hyperparameter configuration. Each entry is the mean $\pm$ standard deviation across seeds 2021--2025. Errors and standard deviations are rounded to three decimals; a reported standard deviation of $0.000$ reflects this rounding.}
\label{tab:rb:shared_seeds}
\begingroup
\scriptsize
\setlength{\tabcolsep}{9pt}
\renewcommand{\arraystretch}{1.04}
\begin{tabular}{lccc}
\toprule
Dataset & Horizon $T$ & MSE & MAE \\
\midrule
ETTh1   & 96    & 0.385 $\pm$ 0.004 & 0.407 $\pm$ 0.002 \\
        & 192   & 0.430 $\pm$ 0.001 & 0.433 $\pm$ 0.001 \\
 & 336          & 0.460 $\pm$ 0.004 & 0.446 $\pm$ 0.003 \\
 & 720          & 0.468 $\pm$ 0.013 & 0.469 $\pm$ 0.007 \\
\midrule
ETTh2 & 96      & 0.333 $\pm$ 0.005 & 0.376 $\pm$ 0.004 \\
 & 192          & 0.409 $\pm$ 0.007 & 0.421 $\pm$ 0.005 \\
 & 336          & 0.405 $\pm$ 0.004 & 0.432 $\pm$ 0.004 \\
 & 720          & 0.442 $\pm$ 0.012 & 0.460 $\pm$ 0.013 \\
\midrule
ETTm1 & 96      & 0.344 $\pm$ 0.006 & 0.373 $\pm$ 0.004 \\
 & 192          & 0.384 $\pm$ 0.003 & 0.392 $\pm$ 0.001 \\
 & 336          & 0.413 $\pm$ 0.007 & 0.415 $\pm$ 0.003 \\
 & 720          & 0.473 $\pm$ 0.006 & 0.450 $\pm$ 0.003 \\
\midrule
ETTm2 & 96      & 0.189 $\pm$ 0.003 & 0.275 $\pm$ 0.002 \\
 & 192          & 0.265 $\pm$ 0.005 & 0.322 $\pm$ 0.002 \\
 & 336          & 0.336 $\pm$ 0.007 & 0.364 $\pm$ 0.003 \\
 & 720          & 0.432 $\pm$ 0.010 & 0.421 $\pm$ 0.005 \\
\midrule
Weather & 96    & 0.172 $\pm$ 0.001 & 0.218 $\pm$ 0.001 \\
 & 192          & 0.217 $\pm$ 0.000 & 0.258 $\pm$ 0.001 \\
 & 336          & 0.273 $\pm$ 0.001 & 0.297 $\pm$ 0.001 \\
 & 720          & 0.346 $\pm$ 0.001 & 0.346 $\pm$ 0.001 \\
\bottomrule
\end{tabular}
\endgroup
\end{table}

\FloatBarrier

\subsection{Six-Dataset Comparison Under Best-Seed Reporting}
\label{app:historical_best_seed}

Table~\ref{tab:best_seed_full} reproduces the earlier six-dataset comparison, reporting \texttt{EntroPE} using its best-seed results. We used an NVIDIA RTX 6000 Ada GPU for the \texttt{EntroPE} runs and took baseline results reported through 2024 from \citet{huang2025timekan}. The main eight-dataset evaluation follows the protocol in Section~\ref{app:main_forecast_protocol}, with per-horizon results in Table~\ref{tab:full_results}.

\begin{sidewaystable}[p]
\centering
\caption{Best-seed forecasting performance (MSE and MAE) of \textbf{EntroPE} and 15 baselines on six datasets. The look-back window is fixed at $L=96$ with forecast horizons $T=\{96,192,336,720\}$. \capbest{Red} marks the best result and \capsecond{Blue} the second best. Avg. is the mean of the four horizons. Experiments are conducted on an NVIDIA RTX Ada 6000 GPU. We extract baseline results up to 2024 from \citet{huang2025timekan}.}
\label{tab:best_seed_full}
\scriptsize
\setlength{\tabcolsep}{2.5pt}
\renewcommand{\arraystretch}{0.95}
\resizebox{\textheight}{!}{%
\begin{tabular}{llcccccccccccccccccccccccccccccccc}
\toprule
\multirow{2}{*}{Dataset} & \multirow{2}{*}{$T$} & \multicolumn{2}{c}{\textbf{EntroPE}} & \multicolumn{2}{c}{TimeMosaic} & \multicolumn{2}{c}{HDMixer} & \multicolumn{2}{c}{TimeBase} & \multicolumn{2}{c}{CALF} & \multicolumn{2}{c}{FilterTS} & \multicolumn{2}{c}{LangTime~\citep{niu2025langtime}} & \multicolumn{2}{c}{TimeMixer} & \multicolumn{2}{c}{Time-FFM} & \multicolumn{2}{c}{TSLANet} & \multicolumn{2}{c}{iTrans.} & \multicolumn{2}{c}{PatchTST} & \multicolumn{2}{c}{TimesNet} & \multicolumn{2}{c}{DLinear} & \multicolumn{2}{c}{FEDformer} & \multicolumn{2}{c}{Autoformer} \\
\cmidrule(lr){3-4}\cmidrule(lr){5-6}\cmidrule(lr){7-8}\cmidrule(lr){9-10}\cmidrule(lr){11-12}\cmidrule(lr){13-14}\cmidrule(lr){15-16}\cmidrule(lr){17-18}\cmidrule(lr){19-20}\cmidrule(lr){21-22}\cmidrule(lr){23-24}\cmidrule(lr){25-26}\cmidrule(lr){27-28}\cmidrule(lr){29-30}\cmidrule(lr){31-32}\cmidrule(lr){33-34}
 & & MSE & MAE & MSE & MAE & MSE & MAE & MSE & MAE & MSE & MAE & MSE & MAE & MSE & MAE & MSE & MAE & MSE & MAE & MSE & MAE & MSE & MAE & MSE & MAE & MSE & MAE & MSE & MAE & MSE & MAE & MSE & MAE \\
\midrule
\multirow{5}{*}{ETTm1} & 96 & \second{0.317} & 0.354 & \best{0.312} & \second{0.351} & 0.340 & 0.369 & 0.373 & 0.388 & 0.319 & \best{0.348} & 0.323 & 0.362 & 0.319 & \best{0.348} & \second{0.317} & 0.356 & 0.336 & 0.369 & 0.321 & 0.362 & 0.334 & 0.368 & 0.352 & 0.374 & 0.338 & 0.375 & 0.346 & 0.374 & 0.379 & 0.419 & 0.505 & 0.475 \\
 & 192 & \best{0.360} & 0.379 & 0.370 & 0.379 & 0.379 & 0.385 & 0.411 & 0.409 & 0.375 & \second{0.376} & 0.364 & 0.382 & 0.368 & \best{0.375} & 0.367 & 0.384 & 0.378 & 0.389 & \second{0.361} & 0.383 & 0.377 & 0.391 & 0.390 & 0.393 & 0.374 & 0.387 & 0.382 & 0.391 & 0.426 & 0.441 & 0.553 & 0.496 \\
 & 336 & 0.388 & \best{0.399} & \second{0.386} & \best{0.399} & 0.398 & 0.409 & 0.436 & 0.421 & 0.410 & \best{0.399} & 0.396 & 0.404 & 0.413 & \second{0.402} & 0.391 & 0.406 & 0.411 & 0.410 & \best{0.382} & 0.404 & 0.426 & 0.420 & 0.421 & 0.414 & 0.410 & 0.411 & 0.415 & 0.415 & 0.445 & 0.459 & 0.621 & 0.537 \\
 & 720 & \best{0.446} & \best{0.433} & 0.469 & 0.440 & 0.467 & 0.443 & 0.503 & 0.461 & 0.478 & 0.439 & 0.462 & 0.441 & 0.487 & 0.439 & \second{0.454} & 0.441 & 0.469 & 0.441 & \best{0.446} & \second{0.438} & 0.491 & 0.459 & 0.462 & 0.449 & 0.478 & 0.450 & 0.473 & 0.451 & 0.543 & 0.490 & 0.671 & 0.561 \\
 & \textbf{Avg.} & \best{0.378} & \best{0.391} & 0.384 & \second{0.392} & 0.396 & 0.402 & 0.431 & 0.420 & 0.396 & \best{0.391} & 0.386 & 0.397 & 0.397 & \best{0.391} & \second{0.382} & 0.397 & 0.399 & 0.402 & \best{0.378} & 0.397 & 0.407 & 0.410 & 0.406 & 0.408 & 0.400 & 0.406 & 0.404 & 0.408 & 0.448 & 0.452 & 0.588 & 0.517 \\
\midrule
\multirow{5}{*}{ETTm2} & 96 & 0.180 & 0.265 & 0.177 & \second{0.254} & 0.183 & 0.266 & 0.188 & 0.271 & \best{0.174} & \best{0.252} & \best{0.174} & 0.256 & 0.188 & 0.258 & \second{0.175} & 0.257 & 0.181 & 0.267 & 0.179 & 0.261 & 0.180 & 0.264 & 0.183 & 0.270 & 0.187 & 0.267 & 0.193 & 0.293 & 0.203 & 0.287 & 0.255 & 0.339 \\
 & 192 & \second{0.241} & 0.311 & \best{0.240} & \best{0.295} & 0.246 & 0.306 & 0.251 & 0.309 & \second{0.241} & \second{0.297} & \best{0.240} & 0.299 & 0.245 & \second{0.297} & \best{0.240} & 0.302 & 0.247 & 0.308 & 0.243 & 0.303 & 0.250 & 0.309 & 0.255 & 0.314 & 0.249 & 0.309 & 0.284 & 0.361 & 0.269 & 0.328 & 0.281 & 0.340 \\
 & 336 & 0.313 & 0.352 & \best{0.298} & \best{0.336} & 0.307 & 0.347 & 0.311 & 0.346 & 0.304 & \second{0.338} & \second{0.301} & 0.339 & \second{0.301} & \best{0.336} & 0.303 & 0.343 & 0.309 & 0.347 & 0.308 & 0.345 & 0.311 & 0.348 & 0.309 & 0.347 & 0.321 & 0.351 & 0.382 & 0.429 & 0.325 & 0.366 & 0.339 & 0.372 \\
 & 720 & 0.411 & 0.413 & \second{0.397} & \best{0.391} & 0.408 & 0.404 & 0.411 & 0.401 & 0.402 & 0.395 & 0.399 & 0.399 & 0.402 & \second{0.393} & \best{0.392} & 0.396 & 0.406 & 0.404 & 0.403 & 0.400 & 0.412 & 0.407 & 0.412 & 0.404 & 0.408 & 0.403 & 0.558 & 0.525 & 0.421 & 0.415 & 0.433 & 0.432 \\
 & \textbf{Avg.} & 0.286 & 0.335 & \best{0.278} & \best{0.319} & 0.286 & 0.331 & 0.290 & 0.332 & 0.280 & \second{0.321} & \second{0.279} & 0.323 & 0.284 & \second{0.321} & \best{0.278} & 0.325 & 0.286 & 0.332 & 0.283 & 0.327 & 0.288 & 0.332 & 0.290 & 0.334 & 0.291 & 0.333 & 0.354 & 0.402 & 0.305 & 0.349 & 0.327 & 0.371 \\
\midrule
\multirow{5}{*}{ETTh1} & 96 & \second{0.375} & 0.399 & \best{0.369} & 0.397 & 0.387 & 0.401 & 0.399 & 0.392 & 0.377 & 0.395 & 0.377 & \second{0.391} & 0.391 & \best{0.388} & 0.385 & 0.402 & 0.385 & 0.400 & 0.387 & 0.405 & 0.386 & 0.405 & 0.460 & 0.447 & 0.384 & 0.402 & 0.397 & 0.412 & 0.395 & 0.424 & 0.449 & 0.459 \\
 & 192 & \second{0.423} & 0.425 & \best{0.419} & 0.439 & 0.441 & 0.428 & 0.455 & \second{0.423} & 0.429 & 0.427 & 0.431 & \second{0.423} & 0.429 & \best{0.419} & 0.443 & 0.430 & 0.439 & 0.430 & 0.448 & 0.436 & 0.441 & 0.436 & 0.512 & 0.477 & 0.436 & 0.429 & 0.446 & 0.441 & 0.469 & 0.470 & 0.500 & 0.482 \\
 & 336 & \best{0.429} & \best{0.432} & 0.454 & 0.451 & 0.452 & \second{0.433} & 0.501 & 0.443 & 0.475 & 0.449 & 0.479 & 0.448 & 0.462 & 0.440 & 0.512 & 0.470 & 0.480 & 0.449 & \second{0.451} & 0.437 & 0.487 & 0.458 & 0.546 & 0.496 & 0.638 & 0.469 & 0.489 & 0.467 & 0.490 & 0.477 & 0.521 & 0.496 \\
 & 720 & \best{0.439} & \second{0.454} & 0.466 & 0.479 & 0.513 & 0.485 & 0.498 & 0.458 & 0.482 & 0.470 & 0.471 & 0.466 & \second{0.458} & \best{0.445} & 0.497 & 0.476 & 0.462 & 0.456 & 0.505 & 0.485 & 0.503 & 0.491 & 0.544 & 0.517 & 0.521 & 0.500 & 0.513 & 0.510 & 0.598 & 0.544 & 0.514 & 0.512 \\
 & \textbf{Avg.} & \best{0.417} & \second{0.428} & \second{0.427} & 0.442 & 0.448 & 0.437 & 0.463 & 0.429 & 0.441 & 0.435 & 0.440 & 0.432 & 0.435 & \best{0.423} & 0.459 & 0.445 & 0.442 & 0.434 & 0.448 & 0.441 & 0.454 & 0.448 & 0.516 & 0.484 & 0.495 & 0.450 & 0.461 & 0.458 & 0.488 & 0.479 & 0.496 & 0.487 \\
\midrule
\multirow{5}{*}{ETTh2} & 96 & \best{0.281} & \best{0.336} & 0.292 & \second{0.341} & 0.289 & \best{0.336} & 0.338 & 0.376 & \second{0.288} & \best{0.336} & 0.293 & 0.344 & 0.299 & \best{0.336} & 0.289 & 0.342 & 0.301 & 0.351 & 0.299 & 0.354 & 0.297 & 0.349 & 0.308 & 0.355 & 0.340 & 0.374 & 0.340 & 0.394 & 0.358 & 0.397 & 0.346 & 0.388 \\
 & 192 & 0.371 & 0.393 & \best{0.367} & 0.391 & 0.386 & 0.397 & 0.402 & 0.405 & \second{0.368} & \second{0.386} & 0.374 & 0.389 & 0.374 & \best{0.382} & 0.378 & 0.397 & 0.378 & 0.397 & 0.382 & 0.399 & 0.380 & 0.400 & 0.393 & 0.405 & 0.402 & 0.414 & 0.482 & 0.479 & 0.429 & 0.439 & 0.456 & 0.452 \\
 & 336 & \second{0.392} & \best{0.394} & 0.417 & 0.427 & 0.438 & 0.442 & 0.437 & 0.440 & 0.415 & 0.423 & 0.411 & 0.423 & 0.410 & 0.418 & 0.432 & 0.434 & 0.422 & 0.431 & \best{0.379} & \second{0.396} & 0.428 & 0.432 & 0.427 & 0.436 & 0.452 & 0.452 & 0.591 & 0.541 & 0.496 & 0.487 & 0.482 & 0.486 \\
 & 720 & 0.421 & \second{0.427} & \best{0.398} & 0.436 & 0.421 & 0.454 & 0.460 & 0.477 & \second{0.417} & 0.435 & 0.423 & 0.441 & 0.418 & \best{0.426} & 0.464 & 0.464 & 0.427 & 0.444 & 0.429 & 0.445 & 0.427 & 0.445 & 0.436 & 0.450 & 0.462 & 0.468 & 0.839 & 0.661 & 0.463 & 0.474 & 0.515 & 0.511 \\
 & \textbf{Avg.} & \best{0.366} & \best{0.388} & \second{0.369} & 0.399 & 0.384 & 0.407 & 0.409 & 0.425 & 0.372 & 0.395 & 0.375 & 0.399 & 0.375 & \second{0.391} & 0.391 & 0.409 & 0.382 & 0.406 & 0.372 & 0.399 & 0.383 & 0.407 & 0.391 & 0.412 & 0.414 & 0.427 & 0.563 & 0.519 & 0.437 & 0.449 & 0.450 & 0.459 \\
\midrule
\multirow{5}{*}{Weather} & 96 & 0.164 & 0.211 & 0.166 & \best{0.201} & 0.175 & 0.223 & 0.170 & 0.215 & 0.164 & 0.205 & \best{0.161} & 0.208 & 0.178 & \second{0.204} & \second{0.163} & 0.209 & 0.191 & 0.230 & 0.177 & 0.217 & 0.174 & 0.214 & 0.186 & 0.227 & 0.172 & 0.220 & 0.195 & 0.252 & 0.217 & 0.296 & 0.266 & 0.336 \\
 & 192 & \best{0.210} & \second{0.252} & 0.219 & 0.258 & 0.226 & 0.265 & 0.216 & 0.256 & 0.213 & \second{0.252} & 0.226 & 0.264 & \second{0.211} & \best{0.249} & \second{0.211} & 0.254 & 0.236 & 0.267 & 0.224 & 0.257 & 0.221 & 0.254 & 0.234 & 0.265 & 0.219 & 0.261 & 0.237 & 0.295 & 0.276 & 0.336 & 0.307 & 0.367 \\
 & 336 & \second{0.256} & \best{0.290} & 0.273 & 0.296 & 0.264 & 0.301 & 0.272 & 0.297 & 0.270 & \second{0.292} & 0.279 & 0.304 & 0.269 & \second{0.292} & 0.263 & 0.293 & 0.289 & 0.303 & 0.278 & 0.297 & 0.278 & 0.296 & 0.284 & 0.301 & \best{0.246} & 0.337 & 0.282 & 0.331 & 0.339 & 0.380 & 0.359 & 0.395 \\
 & 720 & \best{0.339} & \best{0.342} & 0.354 & 0.351 & 0.348 & 0.349 & 0.351 & 0.348 & 0.352 & 0.346 & 0.345 & \second{0.344} & 0.351 & 0.347 & \second{0.344} & 0.348 & 0.362 & 0.350 & 0.355 & 0.347 & 0.358 & 0.347 & 0.356 & 0.349 & 0.365 & 0.359 & 0.345 & 0.382 & 0.403 & 0.428 & 0.419 & 0.428 \\
 & \textbf{Avg.} & \best{0.242} & \second{0.274} & 0.253 & 0.277 & 0.253 & 0.285 & 0.252 & 0.279 & 0.250 & \second{0.274} & 0.253 & 0.280 & 0.252 & \best{0.273} & \second{0.245} & 0.276 & 0.270 & 0.288 & 0.259 & 0.280 & 0.258 & 0.278 & 0.265 & 0.286 & 0.251 & 0.294 & 0.265 & 0.315 & 0.309 & 0.360 & 0.338 & 0.382 \\
\midrule
\multirow{5}{*}{Electricity} & 96 & 0.163 & 0.252 & 0.161 & 0.261 & 0.180 & 0.271 & 0.212 & 0.279 & \best{0.145} & \best{0.239} & 0.153 & 0.247 & 0.181 & 0.266 & 0.153 & 0.245 & 0.198 & 0.282 & 0.175 & 0.260 & \second{0.148} & \second{0.240} & 0.190 & 0.296 & 0.168 & 0.272 & 0.210 & 0.302 & 0.193 & 0.308 & 0.201 & 0.317 \\
 & 192 & 0.177 & 0.268 & 0.176 & 0.274 & 0.184 & 0.275 & 0.209 & 0.281 & \best{0.162} & \best{0.253} & 0.168 & 0.260 & 0.185 & 0.273 & \second{0.166} & \second{0.257} & 0.199 & 0.285 & 0.182 & 0.268 & \best{0.162} & \best{0.253} & 0.199 & 0.304 & 0.184 & 0.322 & 0.210 & 0.305 & 0.201 & 0.315 & 0.222 & 0.334 \\
 & 336 & 0.194 & 0.284 & 0.189 & 0.283 & 0.207 & 0.299 & 0.222 & 0.295 & \best{0.177} & \best{0.268} & 0.187 & 0.278 & 0.198 & 0.281 & 0.185 & 0.275 & 0.212 & 0.298 & 0.199 & 0.285 & \second{0.178} & \second{0.269} & 0.217 & 0.319 & 0.198 & 0.300 & 0.223 & 0.319 & 0.214 & 0.329 & 0.231 & 0.338 \\
 & 720 & 0.235 & 0.321 & 0.225 & \best{0.304} & 0.249 & 0.335 & 0.264 & 0.327 & \second{0.222} & \best{0.304} & 0.227 & 0.313 & 0.241 & 0.320 & 0.224 & \second{0.312} & 0.253 & 0.330 & 0.240 & 0.317 & 0.225 & 0.317 & 0.258 & 0.352 & \best{0.220} & 0.320 & 0.258 & 0.350 & 0.246 & 0.355 & 0.254 & 0.361 \\
 & \textbf{Avg.} & 0.192 & 0.281 & 0.188 & 0.281 & 0.205 & 0.295 & 0.227 & 0.296 & \best{0.177} & \best{0.266} & 0.184 & 0.275 & 0.201 & 0.285 & 0.182 & 0.272 & 0.216 & 0.299 & 0.199 & 0.283 & \second{0.178} & \second{0.270} & 0.216 & 0.318 & 0.193 & 0.304 & 0.225 & 0.319 & 0.214 & 0.327 & 0.227 & 0.338 \\
\bottomrule
\end{tabular}}
\end{sidewaystable}

\FloatBarrier

\section{Computational Cost}
\label{app:efficiency_details}

\subsection{Parameter Counts and Counted Operations}
\label{app:compute_by_dataset}

Table~\ref{tab:efficiency} reports model size and counted multiply--accumulate operations (MACs) per input sample at $L=96$. For $P$ patches of width $d$, global self-attention forms $P^2$ patch-pair scores with cost $\mathcal O(P^2 d)$. The full pipeline also includes entropy estimation, local encoding, and decoder attention. The table specifies which operations are counted.

\begin{table*}[!htbp]
\centering
\caption{Computational cost of all compared models on each dataset, and the average over the eight datasets (Avg.). Parameters are total inference parameters, including frozen weights. MACs are per sample at $L=96$, counted over linear and convolutional layers; attention and FFT products are excluded identically for every model. Each model is instantiated with its own official per-dataset and per-horizon configuration. Lower is better.}
\label{tab:efficiency}
\resizebox{\textwidth}{!}{%
\begin{tabular}{lccccccccc|ccccccccc}
\toprule
\multirow{2}{*}{Models} & \multicolumn{9}{c}{Parameters (M)} & \multicolumn{9}{c}{MACs (G)} \\
\cmidrule(lr){2-10} \cmidrule(lr){11-19}
 & ETTh1 & ETTh2 & ETTm1 & ETTm2 & Weather & Solar & ECL & Traffic & Avg. & ETTh1 & ETTh2 & ETTm1 & ETTm2 & Weather & Solar & ECL & Traffic & Avg. \\
\midrule
FilterTS & 0.546 & 0.565 & 0.703 & 0.369 & 0.246 & 0.438 & 1.032 & 4.909 & 1.101 & 0.005 & 0.006 & 0.007 & 0.003 & 0.006 & 0.038 & 0.273 & 1.178 & 0.190 \\
HDMixer & 0.072 & 0.071 & 0.072 & 0.072 & 0.074 & 0.185 & 0.692 & 4.536 & 0.722 & 0.001 & 0.001 & 0.001 & 0.001 & 0.003 & 0.036 & 0.153 & 0.948 & 0.143 \\
TimeBase & $<$0.001 & $<$0.001 & 0.002 & 0.002 & 0.015 & 0.101 & $<$0.001 & $<$0.001 & 0.015 & $<$0.001 & $<$0.001 & $<$0.001 & $<$0.001 & $<$0.001 & $<$0.001 & 0.001 & 0.004 & 0.001 \\
TimeMosaic & 10.033 & 7.669 & 6.091 & 7.669 & 10.033 & 11.609 & 8.457 & 5.310 & 8.359 & 2.642 & 1.892 & 1.330 & 1.798 & 7.925 & 60.872 & 95.359 & 129.305 & 37.640 \\
TSLANet & 0.172 & 0.172 & 0.620 & 0.342 & 0.235 & 0.235 & 0.715 & 0.715 & 0.401 & 0.003 & 0.003 & 0.023 & 0.004 & 0.009 & 0.058 & 0.868 & 2.330 & 0.412 \\
CALF & 180.439 & 180.439 & 180.439 & 180.439 & 180.439 & 180.439 & 180.439 & 180.439 & 180.439 & 0.680 & 0.680 & 0.680 & 0.680 & 2.039 & 13.299 & 31.161 & 83.679 & 16.612 \\
DLinear & 0.065 & 0.065 & 0.065 & 0.065 & 0.065 & 0.065 & 0.065 & 0.065 & 0.065 & $<$0.001 & $<$0.001 & $<$0.001 & $<$0.001 & 0.001 & 0.009 & 0.021 & 0.056 & 0.011 \\
FEDformer & 16.827 & 16.827 & 16.827 & 16.827 & 16.899 & 17.493 & 18.435 & 21.206 & 17.668 & 2.077 & 2.077 & 2.077 & 2.077 & 2.098 & 2.275 & 2.555 & 3.380 & 2.327 \\
PatchTST & 5.226 & 11.531 & 8.379 & 9.955 & 8.379 & 8.379 & 8.379 & 5.230 & 8.182 & 0.280 & 0.809 & 0.544 & 0.677 & 1.633 & 10.656 & 24.968 & 34.510 & 9.260 \\
TimeMixer & 0.196 & 0.196 & 0.196 & 0.201 & 0.227 & 0.236 & 0.249 & 0.363 & 0.233 & 0.003 & 0.003 & 0.003 & 0.006 & 0.006 & 0.018 & 0.036 & 0.124 & 0.025 \\
iTransformer & 0.255 & 0.255 & 0.255 & 0.255 & 4.957 & 4.957 & 4.957 & 6.535 & 2.803 & 0.003 & 0.003 & 0.003 & 0.003 & 0.124 & 0.699 & 1.610 & 5.656 & 1.012 \\
TimesNet & 0.629 & 1.215 & 2.680 & 1.069 & 1.217 & 1.232 & 150.328 & 301.716 & 57.511 & 1.271 & 2.579 & 3.739 & 1.946 & 2.539 & 2.537 & 324.723 & 648.334 & 123.459 \\
TimeFFM & 100.004 & 100.004 & 100.004 & 100.004 & 100.004 & 100.004 & 100.004 & 100.004 & 100.004 & 5.627 & 5.627 & 5.627 & 5.627 & 16.636 & 107.979 & 252.919 & 678.842 & 134.861 \\
\midrule
\textbf{\oursbold{}} & 0.385 & 0.878 & 0.344 & 0.952 & 0.344 & 0.695 & 0.695 & 1.653 & 0.743 & 0.030 & 0.061 & 0.024 & 0.055 & 0.073 & 0.869 & 1.944 & 12.849 & 1.988 \\
\bottomrule
\end{tabular}}
\end{table*}

\subsection{Model Size in the Longer-Context Comparison}
\label{sec:rb:parameter_counts}

Table~\ref{tab:rb:parameter_counts} compares the medium \texttt{EntroPE} configuration with the comparator sizes reported for each input length and forecast horizon in Section~\ref{sec:rb:long_context}. These ratios depend on the configuration.

\begin{table*}[!htbp]
\centering
\caption{Reported parameter counts for EntroPE-M and comparator models. Parentheses give each comparator's size relative to EntroPE-M. K and M denote thousands and millions of parameters, respectively. Ratios are approximate and configuration-specific.}
\label{tab:rb:parameter_counts}
\begingroup
\scriptsize
\setlength{\tabcolsep}{5pt}
\resizebox{\textwidth}{!}{%
\begin{tabular}{lcccc}
\toprule
$L\!\to\!T$ & EntroPE-M & TimeMosaic & PatchTST & LangTime~\citep{niu2025langtime} \\
\midrule
$36\!\to\!18$   & 176.68K & 6.40M ($36.2\times$) & 408.98K ($2.31\times$) & $\approx$494M+ ($>2796\times$) \\
$512\!\to\!96$  & 952.83K & 9.75M ($10.2\times$) & 1.19M ($1.25\times$)   & $\approx$494M+ ($>518\times$) \\
$512\!\to\!480$ & 4.10M   & 22.49M ($5.5\times$) & 4.33M ($1.06\times$)   & $\approx$494M+ ($>120\times$) \\
\bottomrule
\end{tabular}}
\endgroup
\end{table*}

\subsection{Downstream Training Time Per Epoch}
\label{app:epoch_time}

With boundaries computed during downstream training, the dynamic configuration takes an additional 0.1--1.0 seconds per epoch across the six datasets in Table~\ref{tab:training_time}. These timings cover the downstream epoch. The entropy estimator is trained separately before this stage.

\begin{table}[!htbp]
\centering
\small
\caption{Duration in seconds per downstream training epoch for dynamic and static patching. Overhead percentages are rounded.}
\label{tab:training_time}
\begin{tabular}{lcccccc}
\toprule
\textbf{Metric/Dataset} & \textbf{ETTh1} & \textbf{ETTh2} & \textbf{ETTm1} & \textbf{ETTm2} & \textbf{Weather} & \textbf{ECL} \\
\midrule
Epoch (dynamic) & 3.2 & 3.5 & 17.2 & 20.8 & 54.5 & 181 \\
Epoch (static) & 3.1 & 3.4 & 17.0 & 20.5 & 54.1 & 180 \\
Patching time & 0.1 & 0.1 & 0.2 & 0.3 & 0.4 & 1 \\
Overhead \% & 3\% & 3\% & 1\% & 1\% & 1\% & 1\% \\
\bottomrule
\end{tabular}
\end{table}

\FloatBarrier

\section{Patch Behavior and Robustness}
\label{sec:rb:patch_statistics}

\subsection{Qualitative Examples of Predictive Entropy}
\label{app:qualitative_entropy}

\begin{figure}[!htbp]
    \centering
    \begin{subfigure}{0.86\linewidth}
        \centering
        \includegraphics[width=\linewidth]{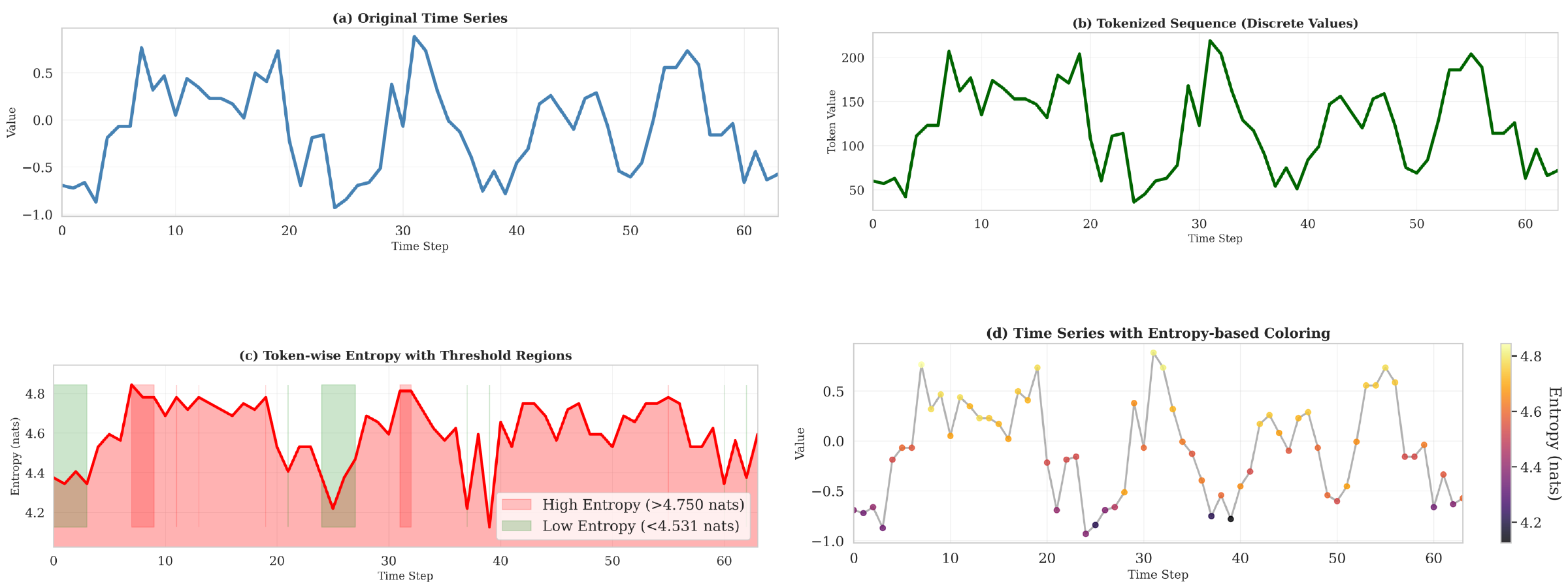}
        \caption{Sample 1}
        \label{fig:enplot1}
    \end{subfigure}

    \begin{subfigure}{0.86\linewidth}
        \centering
        \includegraphics[width=\linewidth]{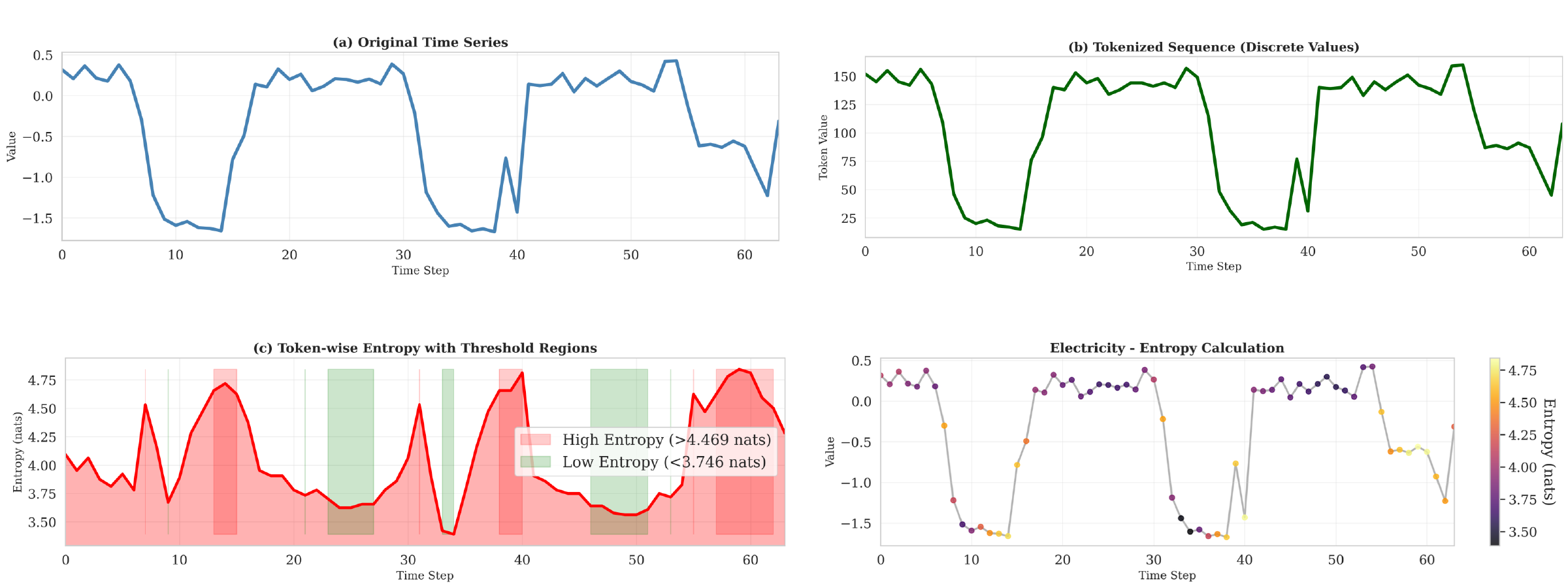}
        \caption{Sample 2}
        \label{fig:enplot2}
    \end{subfigure}

    \begin{subfigure}{0.86\linewidth}
        \centering
        \includegraphics[width=\linewidth]{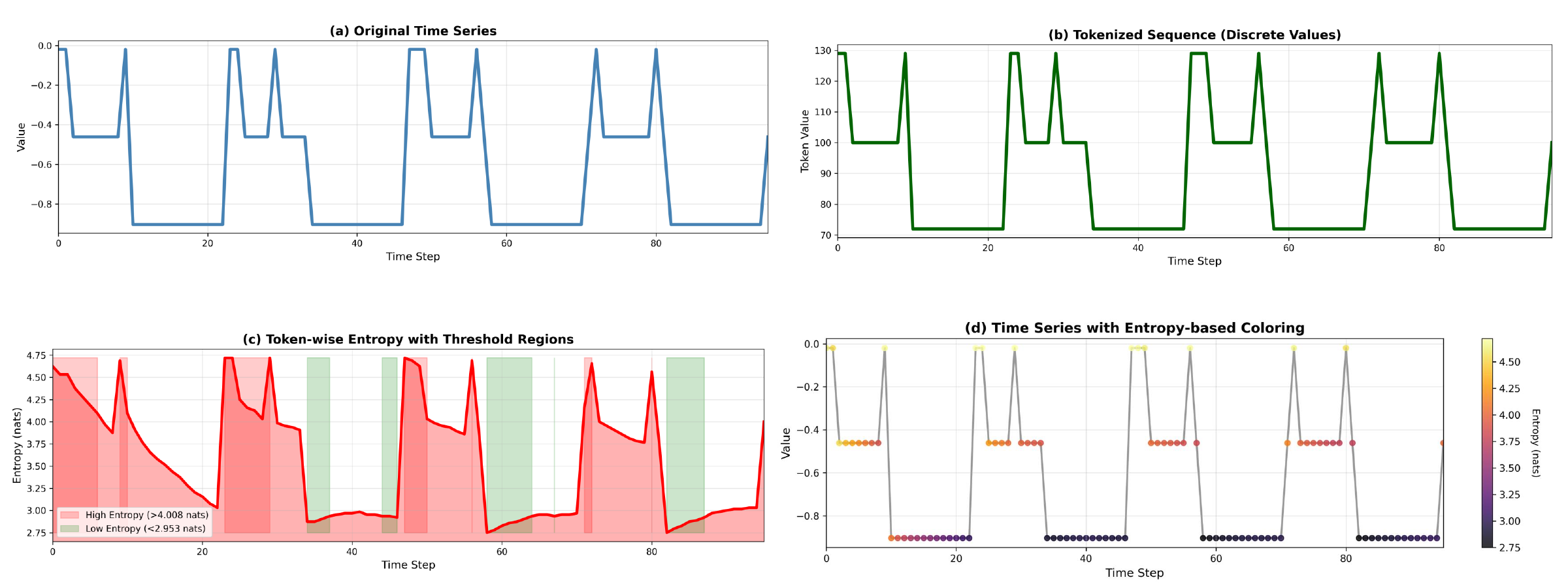}
        \caption{Sample 3}
        \label{fig:enplot3}
    \end{subfigure}

    \caption{Examples from ETT and Electricity. Each example shows the signal, its tokenization, the entropy scores and thresholds, and the signal colored by entropy.}
    \label{fig:entropy_analysis}
\end{figure}

Figure~\ref{fig:entropy_analysis} shows three examples from the observed signal to the entropy score used for segmentation. Within each example, the panels show the normalized series, its discrete tokens, the next-token entropy with threshold regions, and the observed series colored by entropy. In the illustrated ETTh1 window, entropy varies from approximately 3.2 to 4.4 nats. The plots show how the estimator responds within individual windows.

\subsection{Patch Counts and Lengths Across Thresholds}
\label{sec:rb:patch_distributions}

We compute patch statistics on 500 deterministic windows per dataset using $L=512$, seed 2026, and maximum executed patch length 128. Tables~\ref{tab:rb:patch_ett} and~\ref{tab:rb:patch_other} give the mean patch count, mean patch length, and 5th--95th percentile range of patch lengths. Higher thresholds yield fewer, longer patches. We compute segmentation before the downstream encoder, so these statistics are shared across model sizes when the estimator and boundary rule are fixed.

For this diagnostic, 1,235 M4-Monthly series are long enough to provide a 512-observation window. The M4 forecasting results in Table~\ref{tab:rb:long_context} use $L=36$ and $T=18$.

\providecommand{\EntroPERebuttalPatchCell}[3]{%
  \shortstack{#1 / #2\\[1pt]{\scriptsize [1--#3]}}}

\begin{table*}[!htbp]
\centering
\caption{Threshold-dependent patch statistics on the ETT datasets. Each cell gives mean patch count / mean patch length, followed by [P05-P95] patch length. All diagnostics use 500 windows, $L=512$, seed 2026, and maximum executed patch length 128.}
\label{tab:rb:patch_ett}
\begingroup
\tiny
\setlength{\tabcolsep}{7pt}
\renewcommand{\arraystretch}{1.10}
\begin{tabular}{lcccc}
\toprule
Threshold $\alpha$ & ETTh1 & ETTh2 & ETTm1 & ETTm2 \\
\midrule
0.55 & \EntroPERebuttalPatchCell{215.68}{2.37}{9} & \EntroPERebuttalPatchCell{220.61}{2.32}{7} & \EntroPERebuttalPatchCell{224.01}{2.29}{7} & \EntroPERebuttalPatchCell{221.53}{2.31}{7} \\[3pt]
0.60 & \EntroPERebuttalPatchCell{191.67}{2.67}{10} & \EntroPERebuttalPatchCell{195.40}{2.62}{9} & \EntroPERebuttalPatchCell{198.32}{2.58}{9} & \EntroPERebuttalPatchCell{195.95}{2.61}{9} \\[3pt]
0.65 & \EntroPERebuttalPatchCell{164.67}{3.11}{12} & \EntroPERebuttalPatchCell{169.12}{3.03}{11} & \EntroPERebuttalPatchCell{172.46}{2.97}{12} & \EntroPERebuttalPatchCell{170.65}{3.00}{11} \\[3pt]
0.70 & \EntroPERebuttalPatchCell{140.28}{3.65}{14} & \EntroPERebuttalPatchCell{144.52}{3.54}{14} & \EntroPERebuttalPatchCell{147.55}{3.47}{16} & \EntroPERebuttalPatchCell{145.77}{3.51}{14} \\[3pt]
0.75 & \EntroPERebuttalPatchCell{115.67}{4.43}{17} & \EntroPERebuttalPatchCell{119.25}{4.29}{18} & \EntroPERebuttalPatchCell{121.64}{4.21}{22} & \EntroPERebuttalPatchCell{120.42}{4.25}{19} \\[3pt]
0.80 & \EntroPERebuttalPatchCell{92.52}{5.53}{23} & \EntroPERebuttalPatchCell{94.64}{5.41}{23} & \EntroPERebuttalPatchCell{97.57}{5.25}{30} & \EntroPERebuttalPatchCell{96.68}{5.30}{27} \\[3pt]
0.85 & \EntroPERebuttalPatchCell{67.63}{7.57}{37} & \EntroPERebuttalPatchCell{70.49}{7.26}{37} & \EntroPERebuttalPatchCell{72.20}{7.09}{42} & \EntroPERebuttalPatchCell{70.58}{7.25}{42} \\[3pt]
0.90 & \EntroPERebuttalPatchCell{44.21}{11.58}{72} & \EntroPERebuttalPatchCell{46.87}{10.92}{73} & \EntroPERebuttalPatchCell{48.89}{10.47}{67} & \EntroPERebuttalPatchCell{47.93}{10.68}{70} \\[3pt]
0.95 & \EntroPERebuttalPatchCell{22.07}{23.20}{128} & \EntroPERebuttalPatchCell{23.41}{21.87}{124} & \EntroPERebuttalPatchCell{24.62}{20.80}{118} & \EntroPERebuttalPatchCell{24.18}{21.17}{104} \\[3pt]
\bottomrule
\end{tabular}
\endgroup
\end{table*}

\begin{table*}[!htbp]
\centering
\caption{Threshold-dependent patch statistics on Solar, M4-Monthly, and Weather. M4-Monthly uses $L=512$ only for this diagnostic, not for forecasting. Each cell gives mean patch count / mean patch
length, followed by [P05--P95] patch length. All diagnostics use 500 windows, $L=512$, seed 2026, and maximum executed patch length 128.}
\label{tab:rb:patch_other}
\begingroup
\tiny
\setlength{\tabcolsep}{7pt}
\renewcommand{\arraystretch}{1.10}
\begin{tabular}{lccc}
\toprule
Threshold $\alpha$ & Solar & M4-Monthly & Weather \\
\midrule
0.55 & \EntroPERebuttalPatchCell{224.56}{2.28}{3} & \EntroPERebuttalPatchCell{220.44}{2.32}{6} & \EntroPERebuttalPatchCell{226.58}{2.26}{5} \\[3pt]
0.60 & \EntroPERebuttalPatchCell{203.32}{2.52}{2} & \EntroPERebuttalPatchCell{197.83}{2.59}{10} & \EntroPERebuttalPatchCell{201.61}{2.54}{6} \\[3pt]
0.65 & \EntroPERebuttalPatchCell{178.42}{2.87}{4} & \EntroPERebuttalPatchCell{171.53}{2.98}{16} & \EntroPERebuttalPatchCell{175.55}{2.92}{8} \\[3pt]
0.70 & \EntroPERebuttalPatchCell{153.54}{3.33}{6} & \EntroPERebuttalPatchCell{147.35}{3.47}{22} & \EntroPERebuttalPatchCell{151.11}{3.39}{11} \\[3pt]
0.75 & \EntroPERebuttalPatchCell{126.00}{4.06}{9} & \EntroPERebuttalPatchCell{122.10}{4.19}{25} & \EntroPERebuttalPatchCell{125.30}{4.09}{17} \\[3pt]
0.80 & \EntroPERebuttalPatchCell{98.97}{5.17}{18} & \EntroPERebuttalPatchCell{97.84}{5.23}{27} & \EntroPERebuttalPatchCell{100.18}{5.11}{26} \\[3pt]
0.85 & \EntroPERebuttalPatchCell{74.14}{6.91}{43} & \EntroPERebuttalPatchCell{73.62}{6.95}{31} & \EntroPERebuttalPatchCell{75.05}{6.82}{44} \\[3pt]
0.90 & \EntroPERebuttalPatchCell{51.46}{9.95}{101} & \EntroPERebuttalPatchCell{49.82}{10.28}{36} & \EntroPERebuttalPatchCell{50.47}{10.14}{82} \\[3pt]
0.95 & \EntroPERebuttalPatchCell{26.95}{19.00}{128} & \EntroPERebuttalPatchCell{25.16}{20.35}{123} & \EntroPERebuttalPatchCell{26.13}{19.60}{128} \\[3pt]
\bottomrule
\end{tabular}
\endgroup
\end{table*}

\FloatBarrier

\subsection{Sensitivity to Noise and Dataset Transfer}
\label{app:limitations}

\paragraph{Additive noise.}
Table~\ref{tab:noise_boundary_stability} compares ETTh1 boundaries obtained from noisy inputs with those obtained from the clean signal. As the signal-to-noise ratio decreases from 40 to 2 dB, Jaccard similarity falls from 0.626 to 0.281. Mean boundary counts remain close to 22 across these conditions, even as the boundary positions change.

\begin{table}[!htbp]
\centering
\caption{
Boundary stability under additive noise on ETTh1.
Jaccard similarity is measured against clean-signal boundaries.
}
\label{tab:noise_boundary_stability}
\scriptsize
\setlength{\tabcolsep}{5pt}
\renewcommand{\arraystretch}{1}
\begin{tabular}{lcc}
\toprule
\textbf{SNR} & \textbf{Boundary Jaccard} & \textbf{\# Boundaries} \\
\midrule
Clean & 1.000 & 22.35${\pm}$0.00 \\
40 dB & 0.626 & 22.61${\pm}$1.52 \\
20 dB & 0.528 & 22.47${\pm}$1.67 \\
10 dB & 0.419 & 22.27${\pm}$1.71 \\
5 dB  & 0.314 & 21.82${\pm}$2.10 \\
2 dB  & 0.281 & 21.94${\pm}$1.89 \\
\bottomrule
\end{tabular}
\vspace{-1.5mm}
\end{table}

\paragraph{Dataset transfer.}
When we apply the ETTh1-trained entropy estimator to ETTh2 and Weather, the reported boundary Jaccard similarities are 0.34 and 0.25, respectively. Boundary counts remain close to 22. The similar counts and lower overlap show that boundary locations can change under transfer even when segmentation density is preserved.

\section*{Use of Large Language Models}
We refined parts of this paper using a large language model (ChatGPT, GPT-5), exclusively for language polishing and improving clarity of exposition. The authors conceived and executed all ideas, methodology, experiments, and analyses.

\section*{Broader Impacts}
\label{app:broader_impacts}

More accurate forecasts of electricity demand and solar generation could help grid operators plan for changes in supply and demand. In industrial monitoring, \texttt{EntroPE} could also help operators identify sensor patterns that warrant inspection. These applications require care when interpreting patch boundaries. A boundary indicates uncertainty about the next observation, and sensor noise can shift its location even when the underlying equipment is operating normally. Treating each boundary as a fault alarm could therefore trigger unnecessary inspections, while processing channels independently could miss a fault visible only across several sensors. Before deployment, operators should evaluate false alarms and missed events on local data under representative noise and operating conditions, and corroborate alerts before taking consequential action.

\end{document}